\documentclass[12pt]{elsarticle}

\usepackage{amssymb}

\usepackage[utf8]{inputenc}
\usepackage[T1]{fontenc}
\usepackage{amsmath}
\usepackage{graphicx} 
\usepackage[graphicx]{realboxes}
\usepackage{bigstrut}
\usepackage{graphics}
\usepackage{multirow}
\usepackage{caption}
\usepackage{subcaption}
\usepackage{xcolor}
\usepackage{hyperref}

\makeatletter
\def\ps@pprintTitle{%
	\let\@oddhead\@empty
	\let\@evenhead\@empty
	\let\@oddfoot\@empty
	\let\@evenfoot\@oddfoot
}
\makeatother

\newcommand{\B}[1]{\mathbf{#1}}

\begin{document}

\begin{frontmatter}



\title{Meta-Instance Selection. Instance Selection as a Classification Problem with Meta-Features.}


\author[inst1]{Marcin Blachnik 
}
\affiliation[inst1]{organization={Department of Industrial Informatics,\\ Silesian University of Technology},
            addressline={Akademicka 2A}, 
            city={Gliwice},
            postcode={44-100}, 
            country={Poland}}

\author[inst1]{Piotr Ciepliński}



\begin{abstract}
Data pruning, or instance selection, is an important problem in machine learning especially in terms of nearest neighbour classifier. However, in data pruning which speeds up the prediction phase, there is an issue related to the speed and efficiency of the process itself. In response, the study proposes an approach involving transforming the instance selection process into a classification task conducted in a unified meta-feature space where each instance can be classified and assigned to either the "to keep" or "to remove" class. This approach requires training an appropriate meta-classifier, which can be developed based on historical instance selection results from other datasets using reference instance selection methods as a labeling tool. This work proposes constructing the meta-feature space based on properties extracted from the nearest neighbor graph. Experiments conducted on 17 datasets of varying sizes and five reference instance selection methods (ENN, Drop3, ICF, HMN-EI, and CCIS) demonstrate that the proposed solution achieves results comparable to reference instance selection methods while significantly reducing computational complexity.  In the proposed approach, the computational complexity of the system depends only on identifying the k-nearest neighbors for each data sample and running the meta-classifier. Additionally, the study discusses the choice of meta-classifier, recommending the use of Balanced Random Forest.
\end{abstract}




\end{frontmatter}


\section{Intrduction}
Currently, increasing attention is being paid to the issue of filtering data used in the learning process, with a focus on removing useless and noisy data samples. Hence, more and more dedicated solutions are being developed to filter out unimportant samples \cite{sorscher2022beyond} and constitute the new paradigm called data-centric AI\cite{jarrahi2023principles}. Additionally, the scaling law turns into the requirement of huge compute power both during training and prediction which is not always applicable in real live scenarios 
where the compute resources are limited. In that case, both the training data and the prediction model should require small computing resources. Therefore the training set should ensure a possible small size but keep the prediction accuracy of the original training set. This issue is not new, along with its development has been started primarily for the nearest neighbor classifier under the name of instance selection.
Thus, already in the late 1960s and early 1970s, algorithms such as Condensed Nearest Neighbor (CNN), Edited Nearest Neighbor (ENN), and many others were developed. The benchmarks of instance selection indicate the Drop3 \cite{Martinez-ReductionTechniq} and ICF \cite{brighton2002advances} algorithms as the most wildly used, which, despite not being new, are characterized by excellent properties in terms of the balance between the prediction accuracy of the kNN algorithm and the reduction of the size of the stored set of prototypes ($reduction\_rate$) \cite{garcia2016tutorial}. These algorithms are also applicable not only as elements of the learning process for the kNN algorithm (prototype selection as part of the learning process) and hence could also be used as universal algorithms for reducing the size of the training set for any classifier, thereby accelerating the learning process of complex predictive models, the process of finding optimal parameters, etc. Examples of such applications can be found in \cite{blachnik2020comparison,kordos2016reducing} or in \cite{garcia2016tutorial}.

All algorithms belonging to the Instance Selection group share one common element, namely, they explore the structure of the nearest neighbors graph by determining connections between the k nearest samples and their interdependencies to identify vectors that are significant for the decision boundary.

This article is also concerned with the challenge of building an instance selection system, however with a different approach by converting the instance selection process into the binary classification problem. 
Instance selection algorithms typically rely on the analysis of a nearest-neighbor graph. In the nearest neighbor graph, the vertices correspond to individual instances of the training set, and the edges contain information about the nearest neighbors of a given vector. Instance selection algorithms iteratively, often repeatedly, traverse the graph's vertices, allowing to identify whether a particular vector can be important in the learning process or not. But it is important to note, that the characteristics of the nearest neighbors graph (NNG) are common and do not depend on the domain or problem described by the dataset. Therefore, by extracting features that describe the vertices in the NNG the problem of instance selection can be converted into a new meta-dataset in a fixed-sized feature space (meta-features), that is common to all possible datasets. Now, when taking all previously conducted results of any instance selection algorithm on any given dataset the binary information of keeping or rejecting an instance can be used as instance labels, leading to a meta-tanning set. This training set can then be used to train a classifier, which will be called a meta-classifier. In prediction mode, a given input dataset for which the instance selection would be performed is represented as an NGG. Then meta-features are extracted and used as input to the meta-classifier which returns a probability indicating the importance of a given instance. This allows us to select instances without running the original instance selection algorithm.
To maintain satisfactory generalization of the meta-classier, it is significant to create a meta-data set that combines meta-data describing the NNG extracted not from one but multiple datasets, and concatenating all these datasets together.


The conducted analyses show that contrasted to traditional methods of instance selection, for some of them, meta-selection methods allow for better results than the original methods. Moreover, these results are achieved much faster, as the selection process is carried out in a single pass, unlike in instance selection methods, where selection requires iteratively traversing the NNG.

An additional advantage of the developed solution is the description of the training samples' importance as a probability density distribution, indicating how likely a given training vector should be included in the training set without affecting the performance of the classifier or in other words how significant a given vector is in the learning process. The use of a thresholding function allows for determining the number of important samples as a postprocessing step adjusting the desired value to the needs of a given compute resources.

The contribution of this article:
\begin{itemize} 
	\item A new perspective on the data selection process as a classification problem is presented. In the proposed solution a meta-classifier makes decisions about the importance or non-importance of a given training sample.
	\item The development of meta-attributes describing the nearest neighbor graph, which also serve as input to the meta-classifier.
	\item A new metric for instance selection method performance is proposed
\end{itemize}
Article structure: Section \ref{sec:related_work} describes the current state of knowledge in the area of data pruning and meta-learning. Section \ref{sec:meta-is} provides a detailed description of the concept, operation, and construction of the meta-classifier, as well as the feature space used to build the meta-classifier. Section \ref{sec:experiments} presents the experimental design, including the process of building the meta-classifier, performance evauation methods and metrics used to asses its quality. The next section \ref{sec:results} presents the obtained results in comparison to the original instance selection methods. The final chapter \ref{sec:conclusions} contains a summary of the results and future directions for further research.

\section{Related Work} \label{sec:related_work}
In this article, we focus on and combine two areas of machine learning. On the one hand, the article focusses on the problem of training samples selection, in particular on instance selection problem. On the other hand, we modify the concept of meta-learning to develop a meta-classifier used for data pruning.

In data-centric AI, considerable emphasis is placed on assessing data quality and examining the impact of data on the learning process \cite{saha2014data, jain2020overview}. A current overview of the state of knowledge in data-centric AI can be found in \cite{singh2023systematic} and \cite{zha2024data}. According to the proposed taxonomy, data-centric AI can be divided into training data development, inference data development, and data maintenance. Within training data development, an important area is the issue of dataset size reduction, where the goal is to prepare data so that the model can achieve maximum predictive capabilities by removing noisy instances and eliminating redundant data samples that does not impact the quality of predictive models. This cleaning process should aim to ensure maximum prediction accuracy. For example in \cite{sorscher2022beyond}, the authors discuss scaling laws but note that better results can be achieved by appropriately cleaning the dataset. As noted in the introduction, the problem of selecting training data is not new, and over the years, many algorithms have been developed, including instance selection methods. 

The instance selection process was mainly developed for the nearest neighbor classifier but as shown in \cite{blachnik2020comparison,garcia2016tutorial} it can be used also for other classifiers. Instance selection methods focus on removing noisy samples influencing prediction accuracy, and removing redundant samples so that only the minimal subset of samples assuring high prediction accuracy are stored as the final dataset. For many years since 1960 many methods have been developed utilizing many approaches such as focusing on condensation for example as in CNN \cite{Hart-CNN-68} or noise removal such as in ENN \cite{Wilson-ENN-72}. Since this early work, many new methods have been developed. An overview of these methods can be found in \cite{Jankowski04} or in \cite{garcia2012prototype}. This comparison points to Drop3 \cite{Martinez-ReductionTechniq}, ICF \cite{brighton2002advances}, HMN-EI \cite{marchiori2008hit}, and RMHC algorithms (although the last one is a wrapper method that requires an internal classification model). A more recent summary is available in  \cite{cunha2023comparative}, where the authors compared thirteen different algorithms using text classification as a case study. This research points to algorithms based on local sets such as LSSm\cite{leyva2015three} and LSBo\cite{leyva2015three} which were characterized by very high classification accuracy at the cost of size reduction. But also many other algorithms not covered in this comparison were developed. For example in \cite{cunha2023comparative} a method based on ranking has been developed where a special score is used to rank samples. Other modern approaches use the so-called global density-based approach for instance selection. In \cite{malhat2020new} the authors proposed two algorithms based on this approach called global density-based instance selection (GDIS) and its enhanced version called EGDIS. The GDIS algorithm applies a function called relevance function to asses the samples' importance that is based on the purity of k nearest samples and its enhanced version utilized the irrelevance function to improve size reduction. In \cite{blachnik2018ensemble} ensembles of instance selection were proposed where instead of a single method a collection of instance selection methods are evaluated simultaneously and then the obtained results are combined into a single solution. In  \cite{de2019instance} the concept of boosting was applied to improve the quality of selected instances. A very popular approach to instance selection utilizes meta-heuristics such as in \cite{mousavi2020evolutionary} but these methods suffer when the dataset size increases. To overcome this limitation in \cite{blachnik2022fuzzy} a solution was proposed such that the dataset is decomposed into subsets using the conditional clustering method and then within each subset an independent instance of genetic algorithm is used, although this approach was designed to address the regression problem only.

In parallel to the methods related to data-centric AI, meta-learning methods are being developed. These methods focus on utilizing historical knowledge gained from training models on some other datasets, and use this knowledge to build new predictive models. This enables the creation of  models for new datasets faster and with improved predictive accuracy. Additionally, these solutions allow for the identification of models within a set limited time budget to achieve the most efficient model \cite{van2015fast, abdulrahman2018speeding}. In meta-model construction, a commonly used concept is leveraging metadata that describes the characteristics of data sets and reference benchmarks models (so-called landmarks - simple models that can be efficiently trained and evaluated) obtained from datasets on which models were previously trained and whose predictive accuracy is already known. Examples of different meta-features can be found, for instance, in \cite{peng2002improved, castielloetal05, REIF2012, leite2021exploiting, rivolli2022meta}. The simplest solutions in meta-learning involve analyzing the similarity of the examined dataset with historical datasets based on meta-features and creating a corresponding ranking of models and its parameters. Another frequently encountered approach is collaborative filtering \cite{wei2020fast}. Meta-learning methods have also been applied in instance selection \cite{leyva2014use}, although in this case, they have been used to select the most appropriate instance selection method for a given predictive model.

\section{Meta-Classifier-Based Instance Selection} \label{sec:meta-is}
\begin{figure}
	\centering
	\includegraphics[width=0.7\linewidth]{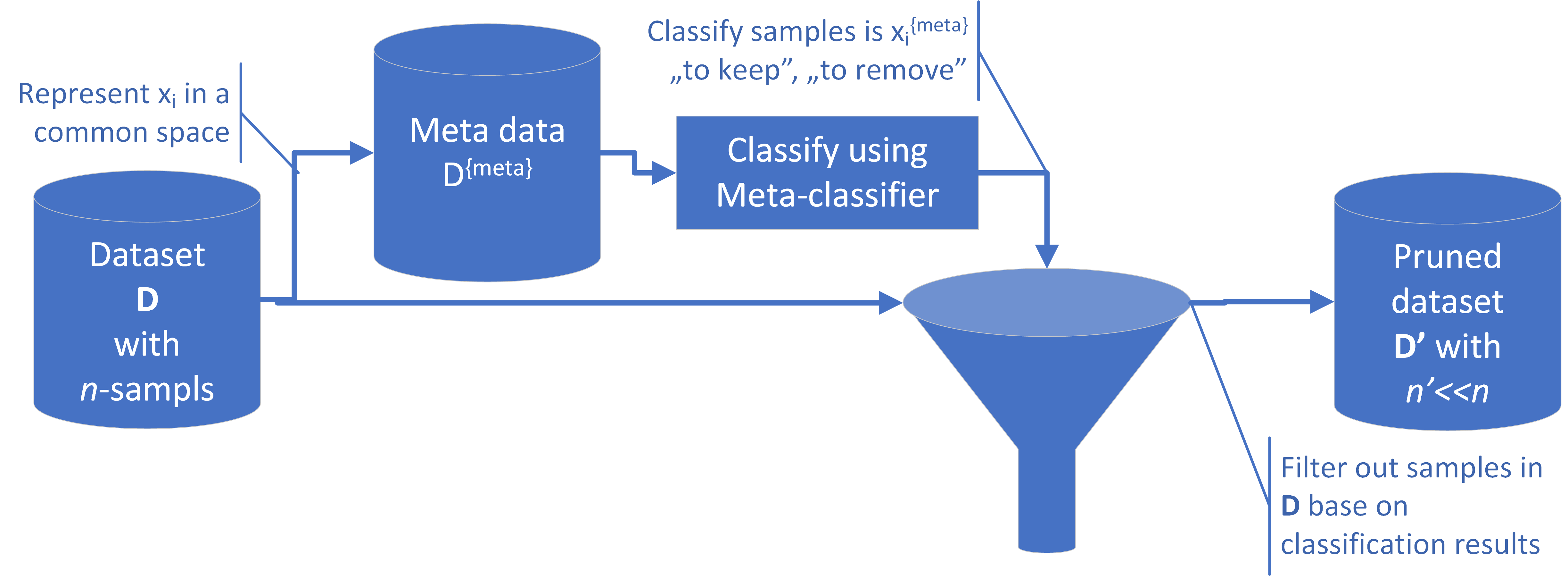}
	\caption{The concept of data processing in the proposed algorithm.}
	\label{fig:basic_idea}
\end{figure}
\begin{figure}
	\centering
	\includegraphics[width=0.5\linewidth]{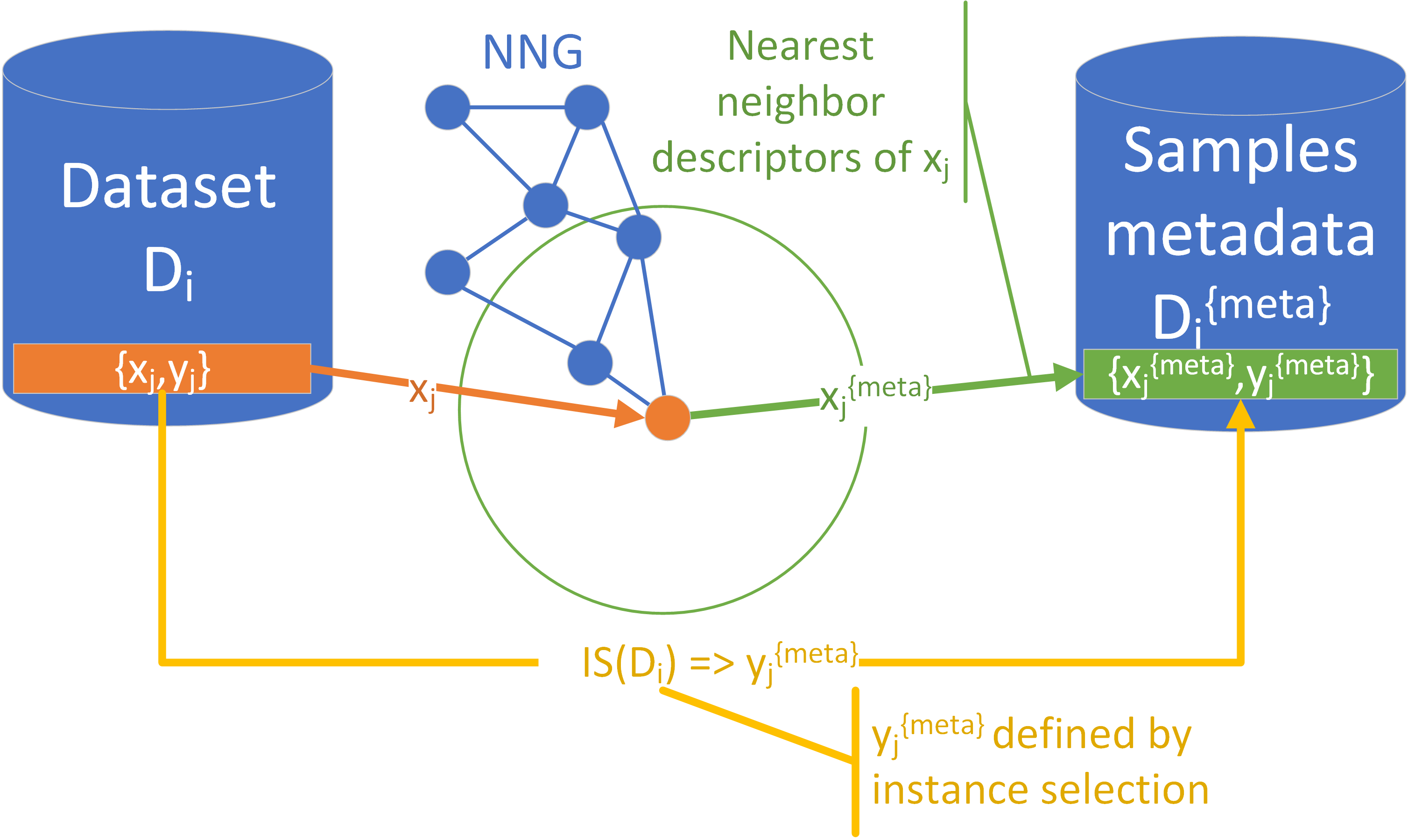}
	\caption{The concept of a single dataset processing in the proposed algorithm. The labelling (marked in yellow) is applied only during the training meta-set preparation.}
	\label{fig:simple_process}
\end{figure}

The proposed algorithm's basic idea is to transform the problem of instance selection into a classification task, where each sample is classified as "to keep" or "to remove" as presented in Figure \ref{fig:basic_idea}. To perform such a task, a classifier and a meta-feature space or instance embeddings are needed in which the classifier will operate.
At the current stage, the meta-space is developed by manually extracting descriptors of the nearest-neighbor graph (NNG), and the meta-classifier is trained on a dataset obtained by combining N independent datasets $\B{D}_i$ each transformed to the meta-space and labeled by the results of the instance selection performed on each of these datasets $\B{D}_i$. 
The basic idea of meta-space extraction is shown in Figure \ref{fig:simple_process}.

Since each of the datasets $\B{D}_i$ is independent each has an independent number of samples $n_i$ so that we can describe it as:
$\B{D}_i = \left\{  
(\B{x}_1,y_1), 
(\B{x}_2,y_2),
\ldots, 
(\B{x}_{n_i},y_{n_i}) 
\right\}$, 
where each pair $(\B{x},y)$, consists of $m_i$ element input vector ($i$ is specific to a given dataset) $\B{x} \in \Re^{m_i}$, and $y \in \left\{ s_1,\ldots,l_{c_i }\right\}$, denotes the label $s$ of the given vector.



Now annotating $\B{G}_i$ as a nearest neighbor graph obtained be $\B{G}_i=NNG(\B{D}_i)$, where the vertices represent individual vectors $\B{x}$, and the edges describe the distances to the nearest neighbors within the dataset  $\B{D}_i$. The function $Meta(\B{G})$ extracts features that describe the properties of the vertices in the graph, converting each vertex into a vector  $\B{x}^{\{meta\}} \in \Re^{n^{\{meta\}}_i}$. By converting each vertex in the NNG, a single meta-dataset is obtained $\B{D}^{\{meta\}}_i$. Hence, it can written as $\B{D}^{\{meta\}}_i  = Meta(NNG(\B{D}_i))$, where both $\B{D}_i$ and $\B{D}^{\{meta\}}_i$ contain the same number of samples. Note that in the meta-set $\B{D}^{\{meta\}}_i$, the number of features  $m^{\{meta\}}_i$ is fixed and doesn't depend on $i$.
Finally, by adopting the notation $IS(\cdot)$ as the function performing instance selection, $\B{D}'_i=IS(\B{D}_i)$ the result is a new dataset containing only those samples which are important to the training process, such that $n_i>n'_i$ and often $n_i>>n'_i$ and the feature space remain unchanged. Then, $\B{D}''_i = \B{D}_i / \B{D}’_i $ contains the irrelevant training samples.

\begin{figure}
	\caption{Graphical representation of the meta-classifier training and application. The green color indicates the procedure for creating the meta-training set, the blue color indicates the procedure for training the meta-classifier, and the orange color indicates the procedure for applying the classifier.}
	\label{fig:procedure_scheme}
	\includegraphics[width=\columnwidth]{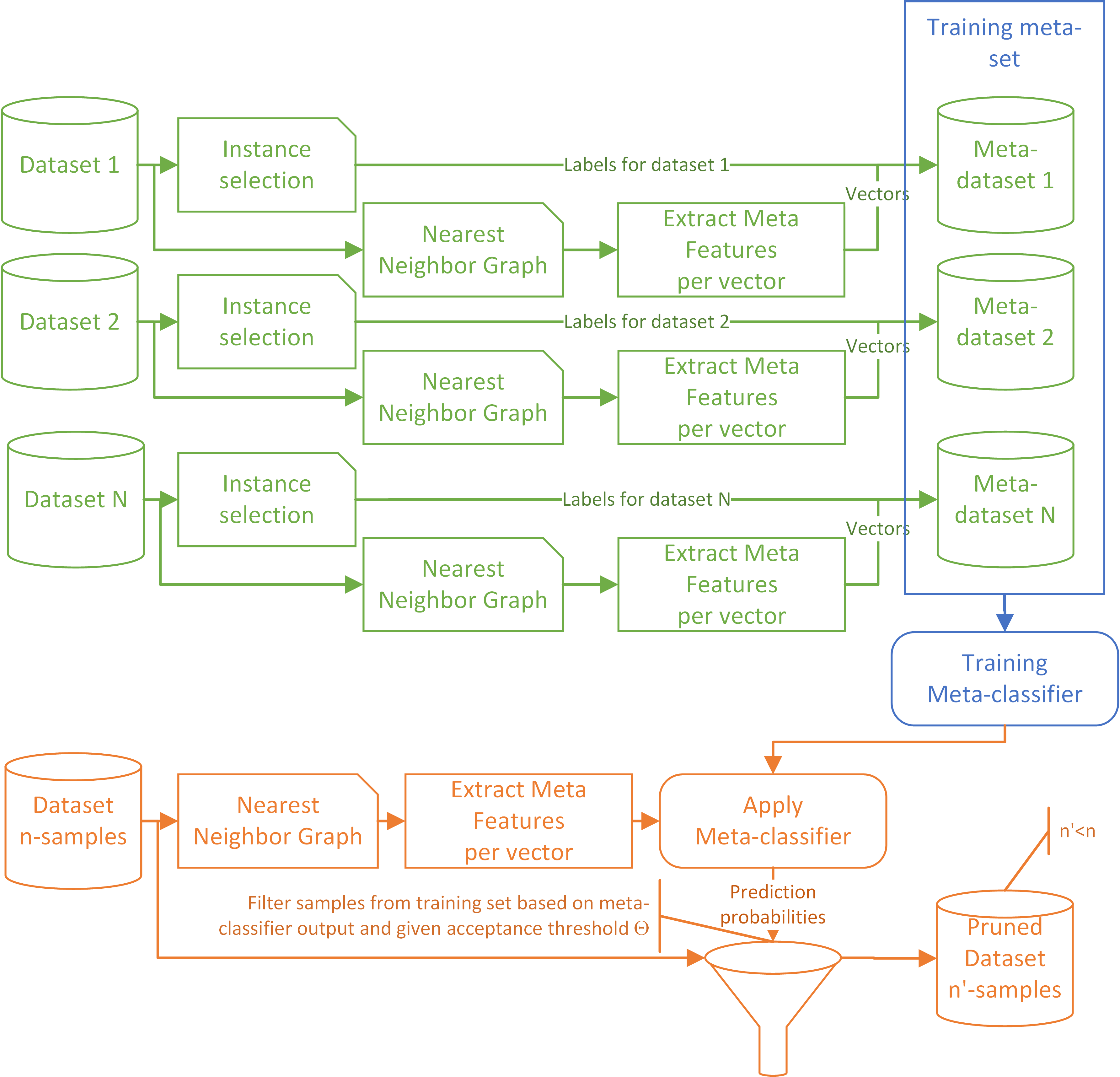}
\end{figure}

By labeling samples from the meta-set $\B{D}^{\{meta\}}_i$ as
\begin{equation*}
	\begin{split}
		\mathop{\forall}\left(\B{x}^{\{meta\}} \in \B{D}^{\{meta\}}_i \right) \text{ if } \B{x}^{\{meta\}} \in \B{D}’_i \\
		\text{ then } y^{\{meta\}} = 1 \\
		\text{ else } y^{\{meta\}} = 0
	\end{split}
\end{equation*}
we get a binary classification problem. 
To train the meta-classifier multiple datasets are combined to achieve proper generalization. More formally each of the data sets $\B{D}_i$ undergoes the same transformation procedure transforming it to $\B{D}^{\{meta\}}_i$, and then each of the $N$ meta-data sets $\B{D}^{\{meta\}}_i$ is merged $\B{D}^{\{meta\}} =\bigcup\limits_{i=1}^N \B{D}^{meta}_i$ into one large meta-data set, on which the meta-model will be trained. The entire procedure of preparing and training the meta-classifier may be written as follows:
\begin{enumerate}
	\item For each data set $\B{D}_i$, determine the NNG $G_i$
	\item From each $G_i$ determine the meta-set $\B{D}^{\{meta\}}_i = Meta(G_i)$ such that $\B{D}_i \rightarrow \B{D}^{\{meta\}}_i$	
	\item For each dataset $\B{D}_i$, perform the selection process of vectors $\B{D}_i’ = IS(\B{D}_i)$ and determine the labels for the meta-dataset $\B{D}^{meta}_i$	
	\item Normalize each of the meta-datasets $Standarize(\B{D}^{\{meta\}}_i)$	
	\item Combine all meta-datasets into one large training meta-dataset
	\item Use $\B{D}^{\{meta\}}$ to train meta-classifier
\end{enumerate}
The procedure for applying the meta-classifier is similar and can be described as:
\begin{enumerate}
	\item For any data set where instance selection $\B{D}_{Te}$ is considered to perform, the dataset should be converted to a meta-set $\B{D}^{\{meta\}}_{Te} = Meta(NNG(\B{D}_{Te}))$	
	\item Make a prediction using the meta-classifier of samples from
	$\B{D}^{\{meta\}}_{Te}$, and as prediction results return the probability of belonging to the class "to be removed" marked as
	$\B{p} = \{p_1,p_2,\ldots,p_{n_{Te}}\}$
	\item Define the threshold $\Theta$ and perform the evaluation assuming the samples labeled as "to be deleted" as the samples for which $p_j<\Theta $, where $p_j $ is the probability that the $j$’th vector of the dataset $\B{D}_{Te}$ is to be deleted	
	\item Remove unnecessary samples from $\B{D}_{Te}$, that is those marked as "to be removed"	
\end{enumerate}
Both procedures are presented graphically in Fig. \ref{fig:procedure_scheme}. The green color indicates the procedure for creating the meta-training set, the blue color indicates the procedure for training the meta-classifier, and the orange color indicates the procedure for applying the meta instance selection process. Note that the training process is executed ones.

\subsection{Meta-Descriptors of the Nearest Neighbor Graph}
The key elements of the algorithm are meta-descriptors or meta-features that describe the local properties of each node of the nearest neighbor graph. 
As suggested by Pinto and others in \cite{pinto2016towards} the process of meta-feature extraction can be represented as a triple - an object, a meta-function, and a post-processing. In the proposed solution, the objects are vertexes of the  NNG, the function is the length of the edges connected to the given vertex of the NNG, and the post-processing is statistics such as mean, minimum, or count of samples fulfilling certain conditions. The condition is determined by the class label of the query vertex and its neighbors. If the labels agree then the edge is marked as the same class, otherwise, it is marked as the opposite.

\begin{itemize}
	\item Average distance to k nearest neighbors from the same class 
	\item Average distance to k nearest neighbors from opposite class
	\item Average distance to k nearest neighbors from any class
	\item Minimum distance to samples from the same class
	\item Minimum distance to samples from the  opposite class
	\item Minimum distance to samples from any class
	\item Number of samples from the same class among k nearest neighbors
	\item Number of samples from the  opposite class among k nearest neighbors
\end{itemize}
If a given value is missing, for instance there are no objects from the  opposite class among k nearest neighbors, the value $-1$ is inserted. In the calculations a squared euclidean distance is used normalized by the number of features. The normalization is essential to make the distance comparable between datasets of different number of features. 

All these features are determined for different values of $k$. Later in the experiments these values were set to $k=\{3, 5, 9, 15, 23, 33\}$.
It is worth noting that from the perspective of computational complexity, it is most effective to conduct the analysis for the maximum value of $k$ (in the given example for $k=33$), and then extract the results for a smaller value of $k$ by ordering the samples according to the distance (the weight of the edges). This procedure allows for a significant reduction in the computational complexity of the algorithm.

The use of different values of $k$ results from the fact that the proposed solution assumes a single pass over the training samples, thus information about the wider surrounding of a given vector for which meta-features are extracted is necessary. It allows for better identification of significance of given instance. For example, when the reference instance selection algorithm that is used for labeling the meta-set uses $k=5$ and it is an iterative algorithm which in the next few iterations of samples pruning identifies that a given vector is redundant, because it lies far from the decision boundary, then similar information for the meta-learning system can be approximated using meta-disscriptors for wide range of $k$ values. In that scenario, the location of objects from the opposite class can be determined by comparing the results for different values of $k$. If for $k={3,5,9,15}$ there are no vectors from the opposite class, for $k=23$ there are few of them, and for $k=33$ almost half of the nearest samples are objects from the opposite class, the algorithm is able to correctly identify the relative position of this vector and decide to remove it or not.

 
\subsection{Computational Complexity Analysis}
Computational complexity can be considered from the perspective of the training process and the application of the developed solution.

The main factor and advantage of the developed solution is the issue of applying the meta instance selection. It consists of two stages. First stage is the procedure of extracting meta-attributes from a new data set $\B{D}_{Te}$ and the second stage is the procedure of applying a meta-classifier. The procedure of extracting meta-attributes depends on the time necessary to create the nearest neighbor graph, which in the worst case requires a computational complexity of $O(n^2)$ (calculating distance between all samples pairs in the dataset $\B{D}_{Te}$ for which instance selection is deliberated to be performed, $|\B{D}_{Te}|=n$). However, using various techniques, e.g. Ball Tree \cite{chubet2023proximity} or KDTree \cite{ram2019revisiting,chen2019fast} algorithms, this complexity can be reduced to $O(n\log(n))$ or using local sensitive hashing \cite{cheng2024robust,jafari2021survey} methods, it is possible to obtain semi-linear complexity. The selection of the method should depend on the number of features in the original data set.

The second stage is the prediction time of the meta-classifier. In this paper, it was decided to use algorithms from the Random Forest family, whose prediction time is $K \log(n^*)$, assuming that the forest consists of $K$ trees and the training set of the meta-classifier has $n^*$ training samples $n^*=\sum\limits_i^N n_i$ (note that the computational complexity depends on the size of the training meta-data set, and not on the size of the dataset for which instance selection is deliberated to be performed). In practice, the maximum depth of the tree is often limited to some fixed number for example to $d=10$, then the complexity is reduced to $K\log(10)$, which can be considered constant from the perspective of the size of the data set $\B{D}_{Te}$. The above indicates that, assuming a constant computational complexity of the classifier, the only  computational cost depending on $n$ is determining the nearest neighbors for each of the training samples, which can be limited to log-linear complexity. Importantly, in comparison to classical instance selection algorithms, the selection procedure is no longer an iterative process but consists in a single pass over the samples from $\B{D}_{Te}$ and a single decision on the importance of a given vector.

An independent element of computational complexity is the time required to prepare and train the meta-classifier. Similarly to the application process, that time depends on the process of extracting meta-descriptors of the nearest neighbour graph and on model training. It is worth underlining that that the meta-classifier training procedure consists of combining the number of samples of each of the $N$ data sets $\B{D}_i$. As a result, the effective size of the training set $D^{\{meta\}}$ consists of $n^*=n_1 + n_2 + \ldots n_N$ training samples. This causes the effective training time to increase radically. Therefore, it was assumed in the studies that the meta-model must be an algorithm that ensures high scalability and low computational complexity of the training process. Such an algorithm is the Random Forest classifier. Additionally it ensures ease of handling imbalanced data by appropriately balancing the size of the sample on which individual trees are trained. As a result, the whole process comes down to complexity $O(N\cdot n^*\log(n^*) + K n^* \log(n^*))$, and the process of meta-descriptor extraction can be easily parallelized by performing the extraction for a single set independently of the other sets in separate processes. Similarly, the Random Forest training process can be easily parallelized, since the individual trees are independent.

\section{System and Experimental Design} \label{sec:experiments}
In order to verify the correctness of the developed algorithm, experiments were conducted experiments comparing the results of several reference instance selection algorithms with the developed solution. The details of the conducted experiments are discussed in the following subsection.

\subsection{Reference methods}
Since the proposed algorithm requires reference methods that will be used for labeling the samples in the meta-set five different reference instance selection algorithms were used in the experiments. These were namely ENN, Drop3, ICF, HMN-EI, and CCIS. These five methods were selected because these are the most often methods used as a reference for all newly developed algorithms. Below a synthetic description of these methods is provided:
\begin{description}
	\item[ENN]
	known as Edited Nearest Neighbor algorithm was developed by Wilson and presented in \cite{Wilson-ENN-72}. It directly addresses problem of regularization and noise removal. It starts by assuming that $\B{D}'=\B{D}$, and in subsequent iterations the algorithm runs over vectors from $\B{D}'$, checking whether most of the nearest neighbors belong to the same category as the tested vector. Although it is a fairly simple and old algorithm dating back to 1972, it is commonly used as a component of other solutions, such as Drop3 or ICF as well as effectively applied in imbalanced learning \cite{JMLR:v18:16-365}.
	\item[Drop]
	is a family of algorithms that is a continuation of the concept proposed in the RNN algorithm \cite{gates1972reduced}. The Drop algorithm \cite{wilson1997instance,Martinez-ReductionTechniq} starts by assuming that initially the selected subset $\B{D}' = \B{D}$, and then each of the instances from $\B{D}'$ is brought to see if its removal will not worsen the prediction accuracy of the kNN classifier. In order to speed up the algorithm, it was proposed to create a nearest neighbours graph storing the k+1 nearest neighbors samples including the so called  enemies (samples with opposite label) and vectors associated with a given training vector (objects whose given vector is one of k nearest neighbors). Such a structure significantly speeds up the evaluation of individual instances. The family starts with Drop1 algorithm which only evaluates individual instances to see if their removal would degrade the results. Drop2 modifies the Drop1 algorithm by the order of instance removal, so that before the main loop begins, the vectors from $\B{D}'$ are sorted in order from the largest to the smallest distance to the nearest enemy. This ensures that vectors far from the decision boundary are removed first. Another modification is the Drop3 algorithm, which starts by running the ENN algorithm to remove outlier vectors and boundary samples, these smoothes the decision boundary. Then, the procedure identical to Drop2 begins. The family also contains Drop4 and Drop5, but since they are not used in the experiments the description would be omitted. Apart from, according to many studies, the Drop3 outperforms the other versions.
	
	\item[ICF]
	stands for \emph{Iterative Case Filtering}, is an algorithm similar to the Drop algorithm with the difference that after analyzing individual instances, the algorithm removes all vectors at once \cite{brighton2002advances}. In this algorithm, the authors define the so-called \emph{LocalSet($\B{x}$)} as a set of vectors belonging to a hypersphere whose center is the examined instance $\B{x}$, and the radius is limited by the nearest instance from the opposite class (excluding this instance). On this basis, two sets are defined, named \emph{Caverage($\B{t}$)} and \emph{Reachable($\B{t}$)}, respectively, as:
	\begin{equation}
		\begin{split}
			Caverage( \B{t}) = \{ \B{x} : \B{x} \in \B{D} \wedge \B{x} \in LocalSet(\B{t}) \} \\
			Reachable(\B{t}) = \{ \B{x} : \B{x} \in \B{D} \wedge \B{t} \in LocalSet(\B{x}) \}
		\end{split}
	\end{equation}
	where $Caverage(\B{t})$ is the set of instances belonging to the set $LocalSet$, whose center is $\B{t}$, or in other words $Caverage(\B{t})$ is a list of nearest neighbors from the same class that $\B{t}$ belongs to, and $Reachable(\B{t})$ is the set of instances from $\B{D}$ for which $\B{t}$ belongs to the $LocalSet$ of these instances, or equivalently, it is a list of vectors associated with $\B{t}$.
	\item[HMN-EI]
	HMN is a family of algorithms, consisting of three solutions called HMN-C, HMN-E and HMN-EI \cite{marchiori2008hit}. The whole is based on Hit Miss Network (HMN), which is a modification of the nearest neighbor graph. In HMN, from each vertex there are $c$ edges to the nearest neighbors from each of the $c$ classes (one nearest neighbor to each category). At the same time, HMN has a counter at each node counting the number of hits and misses, where hits represent the associated vectors for which the class label of the nearest neighbor $y_{nn} = y_x$ ($y_x$the label of vector x), and misses are the scenario in which the nearest neighbor label $y_{nn} \ne y_x$. The individual algorithms HMN-C and HMN-E differ in the rules used to make the decision for removing or leaving instances. In HMN-E four instead of 1 rule used in HMN-C decide whether to remove or leave an instance, the first of which is responsible for removing, and the others for restoring individual instances.	
	The HMN-EI algorithm, runs the HMN-E procedure until the accuracy of the 1-NN classifier no longer deteriorates.
	\item[CCIS]
	CCIS is an algorithm proposed by Elena Marchiori \cite{marchiori2010class} and similarly to algorithms from the HMN family it also uses a hit-miss graph. This algorithm consists of two steps, the first one is called \emph{Class Conditional Selection} (CC) step, while the second one is \emph{Thin-Out Instance Selection} (THIN) step. The CC step is responsible for ensuring proper generalization of the algorithm by removing outliers and boundary vectors. For this purpose, before starting the main loop of the program, candidates to be added to the selected subset $\B{D}'$ are sorted in decreasing order of $Score(a)$ coefficients. Then, the main loop of the CC stage starts, where new candidates are incrementally added to the subset $\B{D}'$ after fulfilling a dedicated condition. The $Score(a)$ function is calculated as the difference:
	\begin{equation}
		Score(a) = K(p_w,p_b)(a) - K(p_b,p_w)(a)
	\end{equation}
	where $K(p_1,p_2)$ is the K-divergence and $p_w(a)$ and $p_b(a)$ are the normalized distributions of the input degree ($inDegree()$) of the within-class and between-class nearest neighbor graph.
\end{description}
All calculations were performed using the implementation of these algorithms created in the Keel project \cite{triguero2017keel} using the Information Selection library \cite{blachnik2016information} available as a plugin for RapidMiner \cite{hofmann2016rapidminer}. They are all implemented using Java environment and in the experiments they were set to use $k=3$ (often used as the default value).

\subsection{The Procedure of Assessing the Quality of Instance Selection}
The development of tests for the meta-instance selection system required the construction of two sub-processes. One in which the meta-classifier is trained and the other in which it is evaluated on a dataset that has not been used to train the meta-classifier. Having in total $N$ data sets, a leave one out test was applied at the datasets level (leave one dataset out). These tests consisted of excluding one of the N available datasets and using N-1 datasets in the process of preparing the meta-classifier, then an instance selection process was carried out on previously excluded dataset. 

During training of the meta-classifier used in meta-instance selection for each of the data sets, firstly the instance selection process was performed using the selected reference algorithm described in the previous section. Secondly, in accordance with the described above procedure, the meta-classifier was trained separately for each of the reference algorithms.

When assessing the quality of instance selection on the single dataset, a 5-fold cross-validation test was performed on that dataset. In this test for each fold, instances were selected using a reference algorithm (ENN, DROP3, HMN-EI, CCIS, ICF), and its performance was assessed using a classifier. Adequately, the meta-instance selection was executed within the same fold, and prediction performance along with the reduction rate was recorded. The performance of instance section methods was assessed using a 1NN  classifier (a typical settings for instance selection), that was trained on the selected samples, and evaluated on the test set for each fold.

However, the meta-instance selection algorithm requires defining an acceptance threshold $\Theta$, therefore, within a single fold, the results were collected for different values of 
$\Theta=\{0.1, 0.2, 0.3, 0.4, 0.5, 0.6, 0.7, 0.8, 0.9\}$. In practice, the calculations were repeated each time for each value of $\Theta$.


In addition to the leave-one-dataset-out test presented in the previous above the additional experiments were performed on larger datasets namely CodrnaNorm and covType. Within these experiments, the meta-classifier was trained on meta-samples extracted from all datasets that were used in the previous experiments and evaluated its performance on the new larger datasets of a size of around 500,000 samples. The size was limited because of the execution time of the reference instance selection methods which, for some datasets took a couple of days. Specifically, for Drop3 after 6 days, the algorithm was stopped without reaching the results, thereby the comparison to Drop3 is not shown within the paper.

\subsection{Datasets Used in the Experiments}
The experiments were performed on data sets available in the Keel project repository \cite{triguero2017keel}. Within this repository, the authors provided data sets of various sizes, from which data sets larger than 2000 samples were selected. Additionally some datasets were obtained from the OpenML project \cite{OpenMLPython2019}. The full characteristics of the datasets are presented in Table \ref{tab:datasets}. There is need to highlight that these data sets are available in the Keel repository with a ready division into cross-validation folds. The division provided in the repository the authors used in the experiments.

\begin{table}[htbp]
	\centering
	\caption{Characteristics of the datasets used in the experiments.}
		\begin{tabular}{|r|l|r|r|r|}
			\hline
			\multicolumn{1}{|l|}{Id} & \textbf{Dataset} & \multicolumn{1}{l|}{\textbf{\# samples}} & \multicolumn{1}{l|}{\textbf{\# attr.}} & \multicolumn{1}{l|}{\textbf{\# classes}} \bigstrut\\
			\hline
			1     & banana & 5300  & 2     & 2 \bigstrut\\
			\hline
			2     & electricity & 45312 & 8     & 2 \bigstrut\\
			\hline
			3     & letter & 20000 & 16    & 26 \bigstrut\\
			\hline
			4     & magic & 19020 & 10    & 2 \bigstrut\\
			\hline
			5     & nursery & 12960 & 8     & 5 \bigstrut\\
			\hline
			6     & optdigits & 5620  & 64    & 10 \bigstrut\\
			\hline
			7     & page-blocks & 5472  & 10    & 5 \bigstrut\\
			\hline
			8    & penbased & 10992 & 16    & 10 \bigstrut\\
			\hline
			9    & phoneme & 5404  & 5     & 2 \bigstrut\\
			\hline
			10    & ring  & 7400  & 20    & 2 \bigstrut\\
			\hline
			11    & satimage & 6435  & 36    & 6 \bigstrut\\
			\hline
			12    & shuttle & 57999 & 9     & 7 \bigstrut\\
			\hline
			13    & spambase & 4597  & 57    & 2 \bigstrut\\
			\hline
			14    & texture & 5500  & 40    & 11 \bigstrut\\
			\hline
			15    & twonorm & 7400  & 20    & 2 \bigstrut\\
			\hline
			16    & codrnaNorm & 488565  & 8    & 2 \bigstrut\\
			\hline
			17    & covType & 581012  & 54   & 2 \bigstrut\\
			\hline
		\end{tabular}%
	\label{tab:datasets}%
\end{table}%

\subsection{Performance Metrics}
To assess the quality of the final classifier, the classical F1 score and the reduction rate were used. Where the reduction rate is defined as: $reduction\_rate=\frac{n-n'}{n}$, where $n$ is the number of samples in the dataset and $n'$ is the number of samples after dataset size reduction.

\begin{figure}
	\centering
	\begin{subfigure}[b]{0.49\textwidth}
		\centering
		\includegraphics[width=\textwidth]{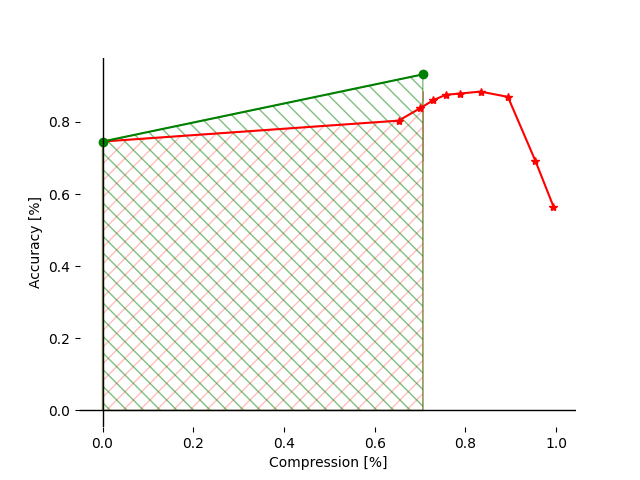}
		\caption{Performance limited}
		\label{fig:perf_1}
	\end{subfigure}
	\hfill
	\begin{subfigure}[b]{0.49\textwidth}
		\centering
		\includegraphics[width=\textwidth]{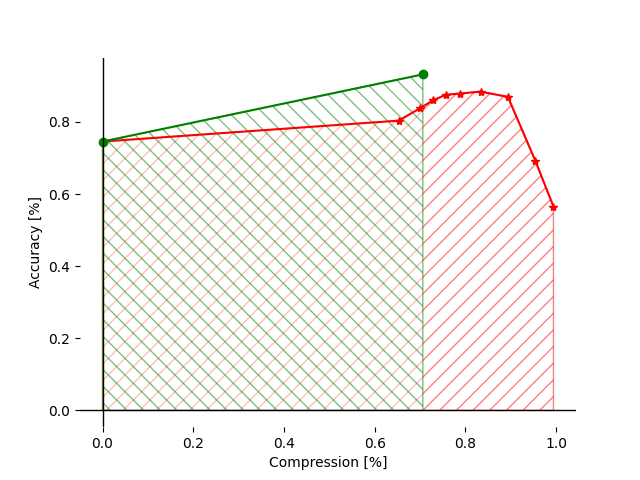}
		\caption{Performance full}
		\label{fig:perf_2}
	\end{subfigure}	
	\caption{Two types of performance measures used in the experiments. The performance is measured as the urea under the accuracy-compression curve. Performance \ref{fig:perf_1} is limited by the reduction rate of the reference instance selection model. Performance \ref{fig:perf_2} is not limited by the reduction rate of the reference instance selection model, it is bounded by the last value of reduction rate. In the figures the area is hatched.}
	\label{fig:perf}
\end{figure}

Since, for the meta-instance selection algorithm it is necessary to define the $\Theta$ threshold, instead of a single result returned by the reference algorithm, for meta instance selection a family of results is obtained. In the experiments the following calculations were performed for $\Theta = \{0.1, 0.2, 0.3, 0.4, 0.5, 0.6, 0.7, 0.8, 0.9\}$. Therefore, for each $\Theta$ a pair of $(prediction\_performance, reduction\_rate)$ was obtained resulting in a series of results. Examples are shown in Figure \ref{fig:perf}.
%
%
This series of results creates a problem when comparing the obtained results with reference instance selection algorithms because the reference methods return a single pair - a single point on the $accuracy-reduction\_rate$ plot. Therefore, in the article, in addition to the typical graphical presentation of the results, they are also evaluated numerically by determining the area under the $accuracy-reduction\_rate$ curve returned by the meta-instance selection algorithm. This area under the $accuracy-reduction\_rate$ also called AUARR is determined by the trapezoidal approximation, starting from the accuracy returned by the model without any instance selection, and then the area is defined by the subsequently collected measurement points obtained for the subsequent values of $\Theta$. This area can be calculated in two ways, one is to limit the maximum $reduction\_rate$ to the $reduction\_rate$ obtained by the reference algorithm (see Figure \ref{fig:perf_1}) - denoted as $AUARR_L$. Then both algorithms are compared assuming that the maximum compression is determined by the reference model. However, this comparison does not reflect the full capabilities of the meta-instance selection algorithm, because with its help it is possible to obtain greater compression, therefore the second of the recorded performance measures is the area under the curve limited only by the last value of the $reduction\_rate$ coefficient denoted as AUARR. In the case of the reference model, this area is determined by linear approximation between the obtained pair (accuracy, $reduction\_rate$) and the accuracy of the 1NN algorithm (Figure \ref{fig:perf_2}).

\subsection{Implementation Details}

The experiments were conducted using the MetaIS library available on GitHub\footnote{\url{https://github.com/mblachnik/MetaIS}}. It implements the developed solution compliant with the API defined by the Scikie-Learn library \cite{scikit-learn} and the Imabalnced-Learn \cite{imbalanced_learn_2017} that extends the basic Scikit-Learn API with the $fit\_resample()$ method where the instance selection algorithm was implemented. The Random Forest -based methods were considered as a meta-classifier. The selection of Random Forest based algorithm was dictated by considering excellent scalability as a function of the size of the training set.  In the proposed solution, the size of the training set increases due to the fact that all partial data sets are glued together. At the same time, due to the high compression of some of the algorithms used in the experiments, the resulting meta-set is unbalanced with the dominance of vectors labeled "to be removed". As a result, the classifiers supporting imbalanced classification was considered as the part of the case study. 

\section{Results} \label{sec:results}
The experiments were divided into 3 parts. The first part compared the results obtained by the meta instance selection with reference models. Then, the time required for the instance selection obtained by the reference algorithm and meta instance selection was compared and analyzed. Eventually the last part shows the influence of model selection on the meta-classifier and its stability.

\subsection{Performance Comparison}
\begin{figure}
	\centering
	\begin{subfigure}[b]{0.27\textwidth}
		\centering
		\includegraphics[width=\textwidth]{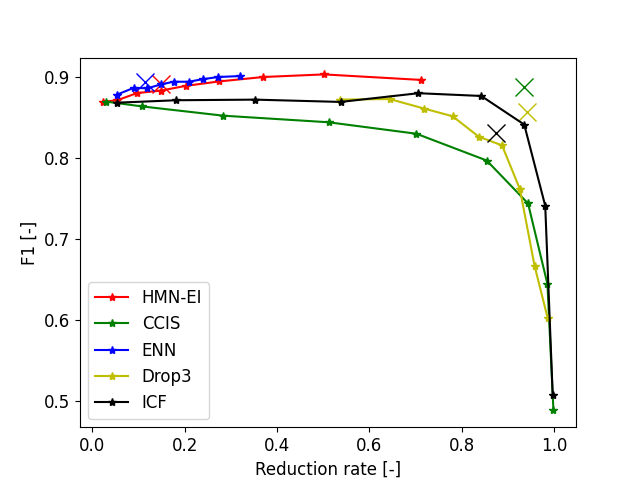}
		\caption{Banana}
		\label{fig:banana}
	\end{subfigure}
	\hfill
	\begin{subfigure}[b]{0.27\textwidth}
		\centering
		\includegraphics[width=\textwidth]{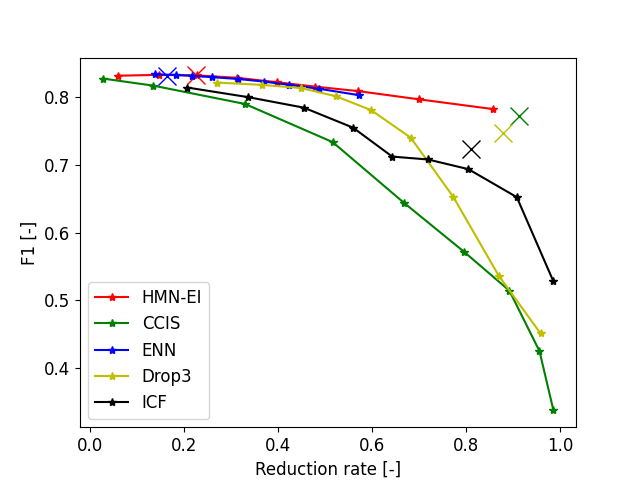}
		\caption{Electricity-norma.}
		\label{fig:electricity}
	\end{subfigure}
	\hfill	
	\begin{subfigure}[b]{0.27\textwidth}
		\centering
		\includegraphics[width=\textwidth]{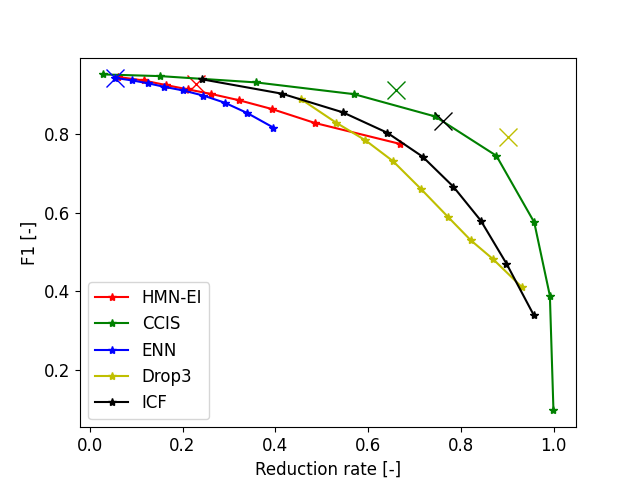}
		\caption{Letter}
		\label{fig:letter}
	\end{subfigure}
	\hfill
	\begin{subfigure}[b]{0.27\textwidth}
		\centering
		\includegraphics[width=\textwidth]{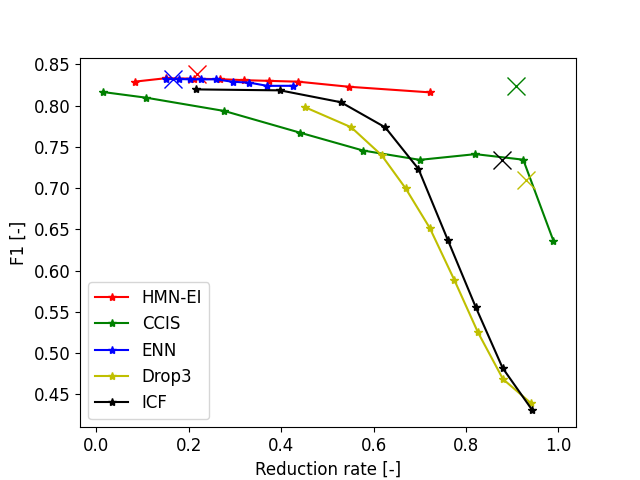}		
		\caption{Magic}
		\label{fig:magic}
	\end{subfigure}
	\hfill	
	\begin{subfigure}[b]{0.27\textwidth}
		\centering
		\includegraphics[width=\textwidth]{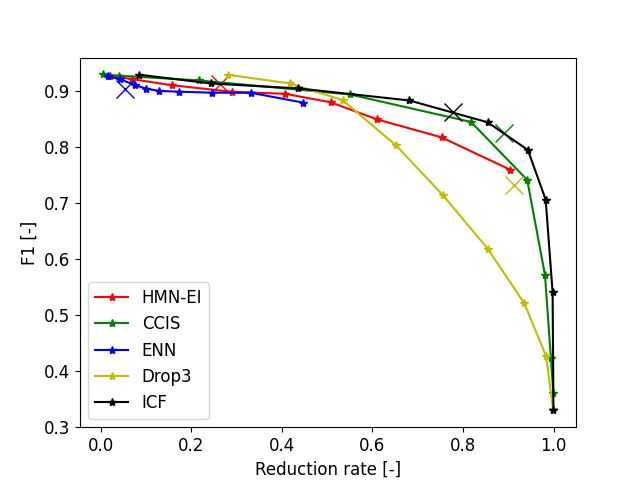}
		\caption{Nursery}
		\label{fig:nursery}
	\end{subfigure}
	\hfill
	\begin{subfigure}[b]{0.27\textwidth}
		\centering
		\includegraphics[width=\textwidth]{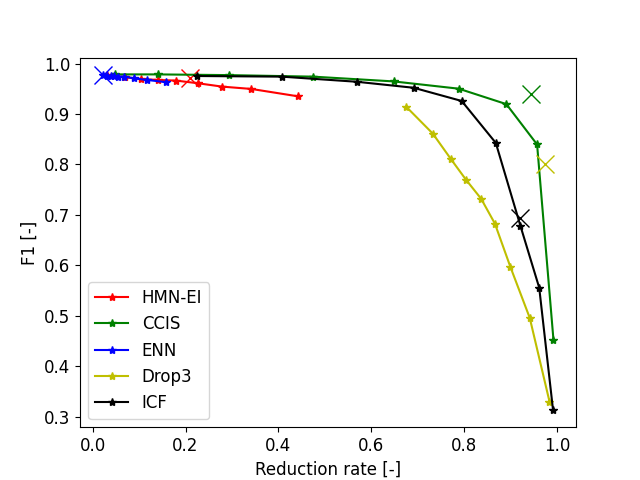}
		\caption{Optidigits}
		\label{fig:optidigits}
	\end{subfigure}
	\hfill
	\begin{subfigure}[b]{0.27\textwidth}
		\centering
		\includegraphics[width=\textwidth]{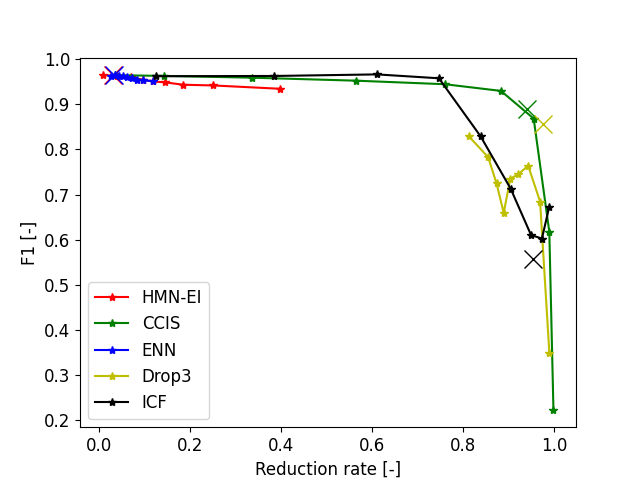}
		\caption{Page-blocks}
		\label{fig:pageblocks}
	\end{subfigure}
	\hfill
	\begin{subfigure}[b]{0.27\textwidth}
		\centering
		\includegraphics[width=\textwidth]{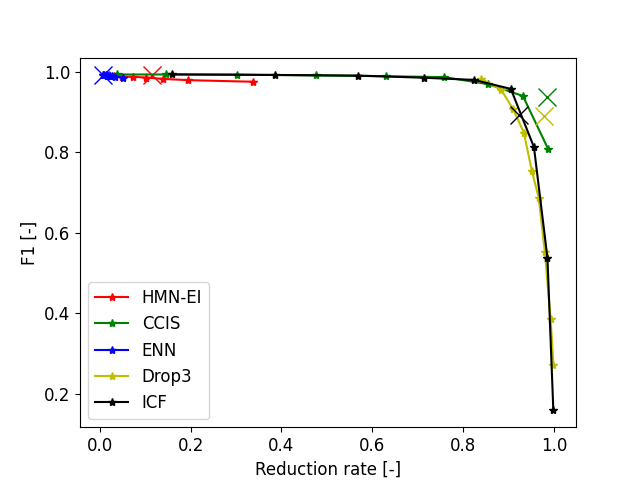}
		\caption{Penbased}
		\label{fig:penbased}
	\end{subfigure}
	\hfill
	\begin{subfigure}[b]{0.27\textwidth}
		\centering
		\includegraphics[width=\textwidth]{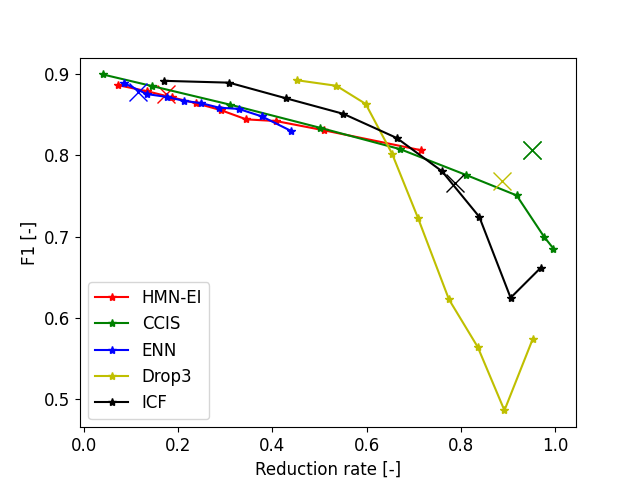}
		\caption{Phoneme}
		\label{fig:phoneme}
	\end{subfigure}
	\hfill
	\begin{subfigure}[b]{0.27\textwidth}
		\centering
		\includegraphics[width=\textwidth]{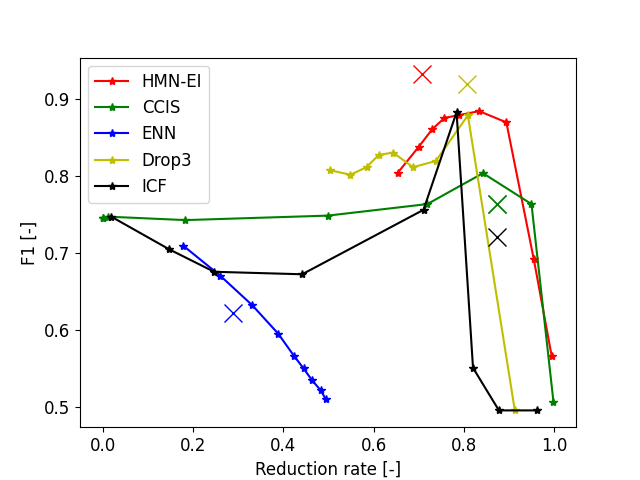}
		\caption{Ring}
		\label{fig:ring}
	\end{subfigure}
	\hfill
	\begin{subfigure}[b]{0.27\textwidth}
		\centering
		\includegraphics[width=\textwidth]{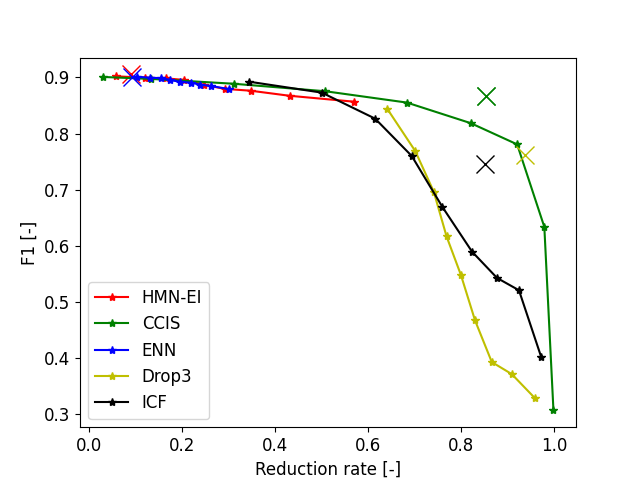}
		\caption{Satimage}
		\label{fig:satimage}
	\end{subfigure}
	\hfill
	\begin{subfigure}[b]{0.27\textwidth}
		\centering
		\includegraphics[width=\textwidth]{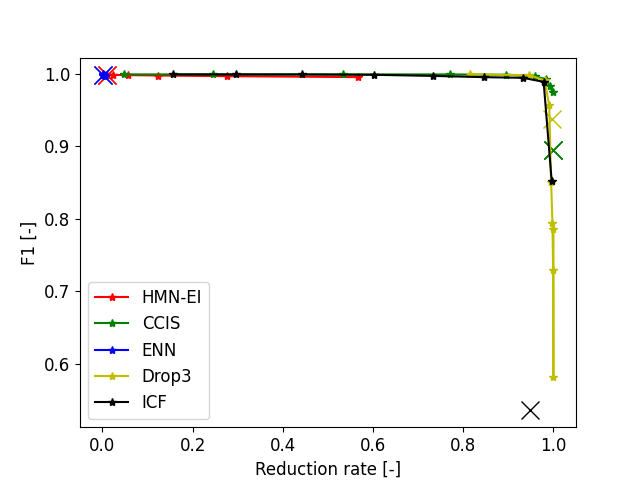}
		\caption{Shuttle}
		\label{fig:shuttle}
	\end{subfigure}
	\hfill
	\begin{subfigure}[b]{0.27\textwidth}
		\centering
		\includegraphics[width=\textwidth]{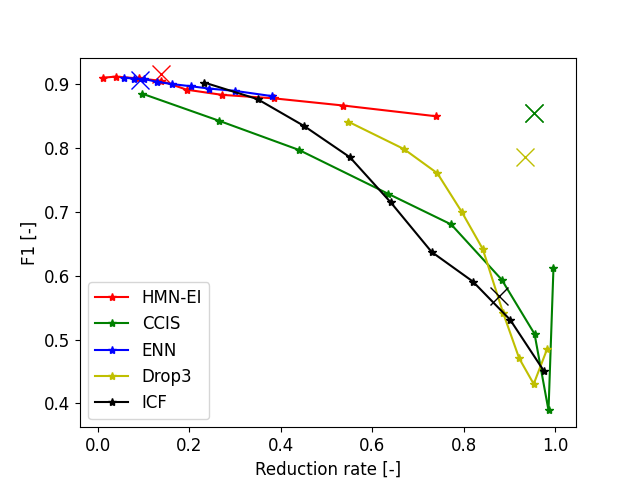}
		\caption{Spambase}
		\label{fig:spambase}
	\end{subfigure}
	\hfill
	\begin{subfigure}[b]{0.27\textwidth}
		\centering
		\includegraphics[width=\textwidth]{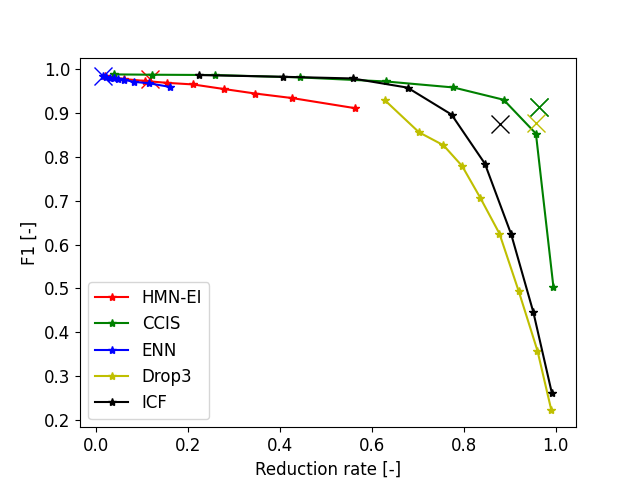}
		\caption{Texture}
		\label{fig:texture}
	\end{subfigure}
	\hfill
	\begin{subfigure}[b]{0.27\textwidth}
		\centering
		\includegraphics[width=\textwidth]{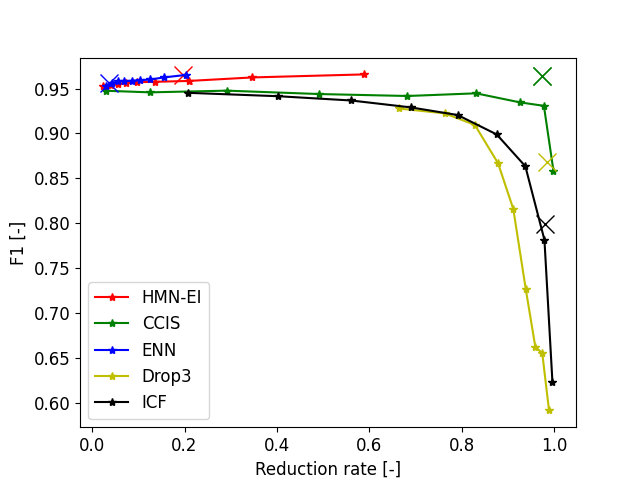}
		\vspace{-1.5\baselineskip}
		\caption{Twonorm}
		\label{fig:twonorm}
	\end{subfigure}
	\vspace{0\baselineskip}
	\caption{Comparison of the reference instance selection methods (marked as X) with the meta instance selection (marked with a curve) in terms of prediction performance and reduction rate. Colors correspond to a particular reference instance selection method and its corresponding meta-instance selection.}
	\label{fig:res_plots}
\end{figure}

Visualization of the performance metrics in particular prediction accuracy and reduction rate are presented in figure \ref{fig:res_plots}. In that figures X indicates given reference method and a curve of the same color corresponds to given meta equivalent to given method. Therefore, in order to properly analyze these figures the results should be analyzed by color comparing X with a curve of the same color. The results are grouped by dataset. This allows to compare not only the relation between compression and prediction performance but also to identify the best instance selection method for a given dataset.
Summarization of these figures by calculating the area under the $accuracy-reduction\_rate$ curve is shown in Tables \ref{tab:results_AUC1d} and \ref{tab:results_AUC}.

A comparison of the results shown in the figures indicates two groups of methods, the first which includes ENN and HMN-EI and the second which contains the three remaining methods (Drop3, ICF, and CCIS). The main difference between these methods appears in compression. The first group has significantly lower reduction rate in comparison to the second group. Therefore these two groups will be analyzed separately.

Within the first group, in all of the cases, there were no differences between the reference ENN and HMN-EI methods marked blue and red respectively with their meta equivalents (Meta-ENN and Meta-HMN-EI). Moreover in almost all of the cases, the use of Meta-ENN and meta-HMN-EI allowed us to achieve significantly higher compression almost without any loss in performance, and even with a little gain. Additionally, the shape of the $accuracy-reduction\_rate$ plot for the meta-instance selection is often flat indicating that even higher compression is possible. The only exception is the Ring data set where Meta-HMN-EI achieved worse accuracy than the reference HMN-EI, but for ENN the accuracy was even higher. 

Within the second group, the $accuracy-reduction\_rate$ curve for the meta-instance selection methods usually indicates a sudden drop after reaching a certain $reduction\_rate$. Before that threshold, the curve is usually flat.  As a result, after a certain $reduction\_rate$ the prediction accuracy starts to decrease rapidly. In those scenarios, the reference instance selection methods often allow achieving higher prediction accuracy for its own $reduction\_rate$, although for a little less $reduction\_rate$ the meta-instance selection methods allow to achieve higher prediction accuracy. That phenomenon appears in Letter, Nursery, Optidigits, Page-blocks, Penbased, Ring, Shuttle Texture, and Twonorm datasets. 

It is also reflected in the results presented in Table \ref{tab:results_AUC1d}, where the area under the $accuracy-reduction\_rate$ (AUARR) is presented, limited by the $reduction\_rate$ of the reference instance selection though. The table presents that, the AUARR indicates the average results of the 5-fold cross-validation obtained for each dataset. The last two rows summarize the table. The row indicated as Wins points to the number of times the MetaIS was significantly better than the reference instance selection method. Here the T-test with $\alpha=0.05$ was used. The last row represents the average difference between the AUARR obtained for meta-instance selection and the reference instance selection method $AUARR_{MetaIS} - AUARR_{RefIS}$. There shall be noted that the positive values indicate that meta instance selection is better and negative values indicate that the reference instance selection method is better. In the brackets are given the results of the Wilcoxon signed-rank test between meta instance selection and reference instance selection methods (here the results over individual datasets were compared). The $(+) $ indicates that the meta instance selection is statistically significantly better ($alpha=0.05$) - here the one-side test was used, and respectively $(-)$ indicates significantly worse results compared to the reference instance selection. The $(=)$ indicates no statistically significant difference. These results indicate that for HMN-EI, ENN, and CCIS there were no statistically significant differences in performance, although, for the HMN-EI method, the MetaHMN-EI won only 3 times, while for the CCIS  the MetaCCIS won 9 times and for ENN the MetaENN won 12 times. For the ICF method, the meta-model significantly outperformed the reference ICF method, while for the Drop3 the meta-model was significantly worse.

\begin{table}[htbp]
	\centering
	\caption{Comparison of the reference instance selection and the meta instance selection equivalence. The values represented the area under the accuracy - $reduction\_rate$ curve ($AUARRC_L$) limited by the compresion of the reference instance selection. The last two lines summarize the results. The one before the last row indicates the number of times MetaIS was significantly better than the reference model. The last row shows the average difference between the performance of MetaIS and the reference instance selection model over all datasets. The symbols (=), (+), and (-) indicate a statistically significant difference between MetaIS and the reference instance selection. (+) means MetaIS is significantly better than IS, and correspondingly (-) that MetaIS is significantly worse.}
	\Rotatebox{90}{
	\resizebox{
		0.8\textheight
		}{!}{	
	\begin{tabular}{|l|c|c|c|c|c|c|c|c|c|c|c|c|c|c|c|c|c|c|c|c|}
		\hline
		\multicolumn{1}{|c|}{\multirow{3}[2]{*}{Dataset}} & \multicolumn{4}{c|}{HMN-EI}   & \multicolumn{4}{c|}{ENN}      & \multicolumn{4}{c}{CCIS}      & \multicolumn{4}{c|}{ICF}      & \multicolumn{4}{c|}{Drop3} \bigstrut[t]\\
		& \multicolumn{2}{c|}{IS} & \multicolumn{2}{c|}{MetaIS} & \multicolumn{2}{c|}{IS} & \multicolumn{2}{c|}{MetaIS} & \multicolumn{2}{c|}{IS} & \multicolumn{2}{c|}{MetaIS} & \multicolumn{2}{c|}{IS} & \multicolumn{2}{c|}{MetaIS} & \multicolumn{2}{c|}{IS} & \multicolumn{2}{c|}{MetaIS} \\
		& \multicolumn{1}{l}{mean} & \multicolumn{1}{l|}{std} & \multicolumn{1}{l}{mean} & \multicolumn{1}{l|}{std} & \multicolumn{1}{l}{mean} & \multicolumn{1}{l|}{std} & \multicolumn{1}{l}{mean} & \multicolumn{1}{l|}{std} & \multicolumn{1}{l}{mean} & \multicolumn{1}{l|}{std} & \multicolumn{1}{l}{mean} & \multicolumn{1}{l|}{std} & \multicolumn{1}{l}{mean} & \multicolumn{1}{l|}{std} & \multicolumn{1}{l}{mean} & \multicolumn{1}{l|}{std} & \multicolumn{1}{l}{mean} & \multicolumn{1}{l|}{std} & \multicolumn{1}{l}{mean} & \multicolumn{1}{l|}{std} \bigstrut[b]\\
		\hline
		banana & \multicolumn{1}{r}{0.1023} & \multicolumn{1}{r|}{0.0007} & \multicolumn{1}{r}{0.1016} & \multicolumn{1}{r|}{0.0008} & \multicolumn{1}{r}{0.0989} & \multicolumn{1}{r|}{0.0009} & \multicolumn{1}{r}{0.0987} & \multicolumn{1}{r|}{0.0006} & \multicolumn{1}{r}{0.8249} & \multicolumn{1}{r|}{0.0058} & \multicolumn{1}{r}{0.7863} & \multicolumn{1}{r|}{0.0091} & \multicolumn{1}{r}{0.7447} & \multicolumn{1}{r|}{0.0066} & \multicolumn{1}{r}{0.7644} & \multicolumn{1}{r|}{0.0073} & \multicolumn{1}{r}{0.8096} & \multicolumn{1}{r|}{0.0084} & \multicolumn{1}{r}{0.8067} & \multicolumn{1}{r|}{0.0101} \bigstrut[t]\\
		electricity-normalized & \multicolumn{1}{r}{0.1877} & \multicolumn{1}{r|}{0.0004} & \multicolumn{1}{r}{0.1880} & \multicolumn{1}{r|}{0.0005} & \multicolumn{1}{r}{0.1364} & \multicolumn{1}{r|}{0.0003} & \multicolumn{1}{r}{0.1367} & \multicolumn{1}{r|}{0.0004} & \multicolumn{1}{r}{0.7316} & \multicolumn{1}{r|}{0.0020} & \multicolumn{1}{r}{0.6553} & \multicolumn{1}{r|}{0.0022} & \multicolumn{1}{r}{0.6270} & \multicolumn{1}{r|}{0.0021} & \multicolumn{1}{r}{0.6267} & \multicolumn{1}{r|}{0.0017} & \multicolumn{1}{r}{0.6904} & \multicolumn{1}{r|}{0.0044} & \multicolumn{1}{r}{0.6767} & \multicolumn{1}{r|}{0.0015} \\
		letter & \multicolumn{1}{r}{0.2169} & \multicolumn{1}{r|}{0.0008} & \multicolumn{1}{r}{0.2155} & \multicolumn{1}{r|}{0.0006} & \multicolumn{1}{r}{0.0501} & \multicolumn{1}{r|}{0.0001} & \multicolumn{1}{r}{0.0502} & \multicolumn{1}{r|}{0.0002} & \multicolumn{1}{r}{0.6189} & \multicolumn{1}{r|}{0.0025} & \multicolumn{1}{r}{0.6150} & \multicolumn{1}{r|}{0.0019} & \multicolumn{1}{r}{0.6814} & \multicolumn{1}{r|}{0.0013} & \multicolumn{1}{r}{0.6737} & \multicolumn{1}{r|}{0.0023} & \multicolumn{1}{r}{0.7861} & \multicolumn{1}{r|}{0.0036} & \multicolumn{1}{r}{0.7253} & \multicolumn{1}{r|}{0.0061} \\
		magic & \multicolumn{1}{r}{0.1809} & \multicolumn{1}{r|}{0.0011} & \multicolumn{1}{r}{0.1812} & \multicolumn{1}{r|}{0.0009} & \multicolumn{1}{r}{0.1350} & \multicolumn{1}{r|}{0.0007} & \multicolumn{1}{r}{0.1351} & \multicolumn{1}{r|}{0.0008} & \multicolumn{1}{r}{0.7460} & \multicolumn{1}{r|}{0.0034} & \multicolumn{1}{r}{0.6997} & \multicolumn{1}{r|}{0.0057} & \multicolumn{1}{r}{0.6821} & \multicolumn{1}{r|}{0.0045} & \multicolumn{1}{r}{0.6713} & \multicolumn{1}{r|}{0.0032} & \multicolumn{1}{r}{0.7079} & \multicolumn{1}{r|}{0.0082} & \multicolumn{1}{r}{0.6756} & \multicolumn{1}{r|}{0.0010} \\
		marketing & \multicolumn{1}{r}{0.1438} & \multicolumn{1}{r|}{0.0037} & \multicolumn{1}{r}{0.1427} & \multicolumn{1}{r|}{0.0036} & \multicolumn{1}{r}{0.2068} & \multicolumn{1}{r|}{0.0035} & \multicolumn{1}{r}{0.2042} & \multicolumn{1}{r|}{0.0018} & \multicolumn{1}{r}{0.0574} & \multicolumn{1}{r|}{0.0015} & \multicolumn{1}{r}{0.0496} & \multicolumn{1}{r|}{0.0016} & \multicolumn{1}{r}{0.2295} & \multicolumn{1}{r|}{0.0023} & \multicolumn{1}{r}{0.2362} & \multicolumn{1}{r|}{0.0058} & \multicolumn{1}{r}{0.2365} & \multicolumn{1}{r|}{0.0029} & \multicolumn{1}{r}{0.2432} & \multicolumn{1}{r|}{0.0052} \\
		nursery & \multicolumn{1}{r}{0.2514} & \multicolumn{1}{r|}{0.0041} & \multicolumn{1}{r}{0.2493} & \multicolumn{1}{r|}{0.0056} & \multicolumn{1}{r}{0.0626} & \multicolumn{1}{r|}{0.0014} & \multicolumn{1}{r}{0.0630} & \multicolumn{1}{r|}{0.0015} & \multicolumn{1}{r}{0.7332} & \multicolumn{1}{r|}{0.0110} & \multicolumn{1}{r}{0.7483} & \multicolumn{1}{r|}{0.0102} & \multicolumn{1}{r}{0.7206} & \multicolumn{1}{r|}{0.0103} & \multicolumn{1}{r}{0.7263} & \multicolumn{1}{r|}{0.0132} & \multicolumn{1}{r}{0.7708} & \multicolumn{1}{r|}{0.0234} & \multicolumn{1}{r}{0.7804} & \multicolumn{1}{r|}{0.0095} \\
		optdigits & \multicolumn{1}{r}{0.2029} & \multicolumn{1}{r|}{0.0008} & \multicolumn{1}{r}{0.2021} & \multicolumn{1}{r|}{0.0006} & \multicolumn{1}{r}{0.0222} & \multicolumn{1}{r|}{0.0001} & \multicolumn{1}{r}{0.0222} & \multicolumn{1}{r|}{0.0001} & \multicolumn{1}{r}{0.9078} & \multicolumn{1}{r|}{0.0051} & \multicolumn{1}{r}{0.9112} & \multicolumn{1}{r|}{0.0050} & \multicolumn{1}{r}{0.7681} & \multicolumn{1}{r|}{0.0087} & \multicolumn{1}{r}{0.8719} & \multicolumn{1}{r|}{0.0038} & \multicolumn{1}{r}{0.8650} & \multicolumn{1}{r|}{0.0166} & \multicolumn{1}{r}{0.8509} & \multicolumn{1}{r|}{0.0046} \\
		page-blocks & \multicolumn{1}{r}{0.0317} & \multicolumn{1}{r|}{0.0002} & \multicolumn{1}{r}{0.0317} & \multicolumn{1}{r|}{0.0002} & \multicolumn{1}{r}{0.0321} & \multicolumn{1}{r|}{0.0002} & \multicolumn{1}{r}{0.0321} & \multicolumn{1}{r|}{0.0002} & \multicolumn{1}{r}{0.8691} & \multicolumn{1}{r|}{0.0086} & \multicolumn{1}{r}{0.8929} & \multicolumn{1}{r|}{0.0063} & \multicolumn{1}{r}{0.7205} & \multicolumn{1}{r|}{0.0231} & \multicolumn{1}{r}{0.8807} & \multicolumn{1}{r|}{0.0074} & \multicolumn{1}{r}{0.8849} & \multicolumn{1}{r|}{0.0200} & \multicolumn{1}{r}{0.8480} & \multicolumn{1}{r|}{0.0186} \\
		penbased & \multicolumn{1}{r}{0.1142} & \multicolumn{1}{r|}{0.0002} & \multicolumn{1}{r}{0.1138} & \multicolumn{1}{r|}{0.0003} & \multicolumn{1}{r}{0.0070} & \multicolumn{1}{r|}{0.0000} & \multicolumn{1}{r}{0.0070} & \multicolumn{1}{r|}{0.0000} & \multicolumn{1}{r}{0.9497} & \multicolumn{1}{r|}{0.0051} & \multicolumn{1}{r}{0.9658} & \multicolumn{1}{r|}{0.0014} & \multicolumn{1}{r}{0.8761} & \multicolumn{1}{r|}{0.0015} & \multicolumn{1}{r}{0.9165} & \multicolumn{1}{r|}{0.0017} & \multicolumn{1}{r}{0.9212} & \multicolumn{1}{r|}{0.0075} & \multicolumn{1}{r}{0.9500} & \multicolumn{1}{r|}{0.0041} \\
		phoneme & \multicolumn{1}{r}{0.1521} & \multicolumn{1}{r|}{0.0013} & \multicolumn{1}{r}{0.1517} & \multicolumn{1}{r|}{0.0014} & \multicolumn{1}{r}{0.1007} & \multicolumn{1}{r|}{0.0010} & \multicolumn{1}{r}{0.1010} & \multicolumn{1}{r|}{0.0009} & \multicolumn{1}{r}{0.8142} & \multicolumn{1}{r|}{0.0048} & \multicolumn{1}{r}{0.7940} & \multicolumn{1}{r|}{0.0069} & \multicolumn{1}{r}{0.6592} & \multicolumn{1}{r|}{0.0071} & \multicolumn{1}{r}{0.6818} & \multicolumn{1}{r|}{0.0071} & \multicolumn{1}{r}{0.7458} & \multicolumn{1}{r|}{0.0114} & \multicolumn{1}{r}{0.7336} & \multicolumn{1}{r|}{0.0038} \\
		ring  & \multicolumn{1}{r}{0.5909} & \multicolumn{1}{r|}{0.0018} & \multicolumn{1}{r}{0.5482} & \multicolumn{1}{r|}{0.0045} & \multicolumn{1}{r}{0.2007} & \multicolumn{1}{r|}{0.0010} & \multicolumn{1}{r}{0.2083} & \multicolumn{1}{r|}{0.0011} & \multicolumn{1}{r}{0.6594} & \multicolumn{1}{r|}{0.0024} & \multicolumn{1}{r}{0.6600} & \multicolumn{1}{r|}{0.0069} & \multicolumn{1}{r}{0.6426} & \multicolumn{1}{r|}{0.0052} & \multicolumn{1}{r}{0.6143} & \multicolumn{1}{r|}{0.0096} & \multicolumn{1}{r}{0.6722} & \multicolumn{1}{r|}{0.0046} & \multicolumn{1}{r}{0.6412} & \multicolumn{1}{r|}{0.0041} \\
		satimage & \multicolumn{1}{r}{0.0831} & \multicolumn{1}{r|}{0.0003} & \multicolumn{1}{r}{0.0829} & \multicolumn{1}{r|}{0.0002} & \multicolumn{1}{r}{0.0859} & \multicolumn{1}{r|}{0.0004} & \multicolumn{1}{r}{0.0859} & \multicolumn{1}{r|}{0.0005} & \multicolumn{1}{r}{0.7559} & \multicolumn{1}{r|}{0.0039} & \multicolumn{1}{r}{0.7459} & \multicolumn{1}{r|}{0.0035} & \multicolumn{1}{r}{0.7047} & \multicolumn{1}{r|}{0.0075} & \multicolumn{1}{r}{0.7123} & \multicolumn{1}{r|}{0.0073} & \multicolumn{1}{r}{0.7812} & \multicolumn{1}{r|}{0.0260} & \multicolumn{1}{r}{0.7319} & \multicolumn{1}{r|}{0.0049} \\
		shuttle & \multicolumn{1}{r}{0.0111} & \multicolumn{1}{r|}{0.0000} & \multicolumn{1}{r}{0.0111} & \multicolumn{1}{r|}{0.0000} & \multicolumn{1}{r}{0.0011} & \multicolumn{1}{r|}{0.0000} & \multicolumn{1}{r}{0.0011} & \multicolumn{1}{r|}{0.0000} & \multicolumn{1}{r}{0.9449} & \multicolumn{1}{r|}{0.0158} & \multicolumn{1}{r}{0.9966} & \multicolumn{1}{r|}{0.0004} & \multicolumn{1}{r}{0.7261} & \multicolumn{1}{r|}{0.0198} & \multicolumn{1}{r}{0.9441} & \multicolumn{1}{r|}{0.0004} & \multicolumn{1}{r}{0.9652} & \multicolumn{1}{r|}{0.0239} & \multicolumn{1}{r}{0.9944} & \multicolumn{1}{r|}{0.0005} \\
		spambase & \multicolumn{1}{r}{0.1261} & \multicolumn{1}{r|}{0.0007} & \multicolumn{1}{r}{0.1256} & \multicolumn{1}{r|}{0.0007} & \multicolumn{1}{r}{0.0862} & \multicolumn{1}{r|}{0.0003} & \multicolumn{1}{r}{0.0863} & \multicolumn{1}{r|}{0.0004} & \multicolumn{1}{r}{0.8445} & \multicolumn{1}{r|}{0.0080} & \multicolumn{1}{r}{0.7326} & \multicolumn{1}{r|}{0.0079} & \multicolumn{1}{r}{0.6507} & \multicolumn{1}{r|}{0.0165} & \multicolumn{1}{r}{0.7001} & \multicolumn{1}{r|}{0.0084} & \multicolumn{1}{r}{0.7917} & \multicolumn{1}{r|}{0.0195} & \multicolumn{1}{r}{0.7559} & \multicolumn{1}{r|}{0.0064} \\
		texture & \multicolumn{1}{r}{0.1137} & \multicolumn{1}{r|}{0.0005} & \multicolumn{1}{r}{0.1133} & \multicolumn{1}{r|}{0.0004} & \multicolumn{1}{r}{0.0139} & \multicolumn{1}{r|}{0.0001} & \multicolumn{1}{r}{0.0139} & \multicolumn{1}{r|}{0.0001} & \multicolumn{1}{r}{0.9176} & \multicolumn{1}{r|}{0.0081} & \multicolumn{1}{r}{0.9340} & \multicolumn{1}{r|}{0.0038} & \multicolumn{1}{r}{0.8123} & \multicolumn{1}{r|}{0.0017} & \multicolumn{1}{r}{0.8348} & \multicolumn{1}{r|}{0.0044} & \multicolumn{1}{r}{0.8945} & \multicolumn{1}{r|}{0.0099} & \multicolumn{1}{r}{0.8433} & \multicolumn{1}{r|}{0.0123} \\
		twonorm & \multicolumn{1}{r}{0.1354} & \multicolumn{1}{r|}{0.0007} & \multicolumn{1}{r}{0.1352} & \multicolumn{1}{r|}{0.0007} & \multicolumn{1}{r}{0.0372} & \multicolumn{1}{r|}{0.0002} & \multicolumn{1}{r}{0.0371} & \multicolumn{1}{r|}{0.0002} & \multicolumn{1}{r}{0.9295} & \multicolumn{1}{r|}{0.0033} & \multicolumn{1}{r}{0.9177} & \multicolumn{1}{r|}{0.0041} & \multicolumn{1}{r}{0.8607} & \multicolumn{1}{r|}{0.0158} & \multicolumn{1}{r}{0.9132} & \multicolumn{1}{r|}{0.0037} & \multicolumn{1}{r}{0.8920} & \multicolumn{1}{r|}{0.0056} & \multicolumn{1}{r}{0.8969} & \multicolumn{1}{r|}{0.0029} \\
		codrnaNorm & \multicolumn{1}{r}{0.1296} & \multicolumn{1}{r|}{0.0001} & \multicolumn{1}{r}{0.1289} & \multicolumn{1}{r|}{0.0001} & \multicolumn{1}{r}{0.0459} & \multicolumn{1}{r|}{0.0000} & \multicolumn{1}{r}{0.0459} & \multicolumn{1}{r|}{0.0000} & \multicolumn{1}{r}{0.9303} & \multicolumn{1}{r|}{0.0010} & \multicolumn{1}{r}{0.9341} & \multicolumn{1}{r|}{0.0006} & \multicolumn{1}{r}{0.6869} & \multicolumn{1}{r|}{0.0041} & \multicolumn{1}{r}{0.9177} & \multicolumn{1}{r|}{0.0011} & \multicolumn{1}{r}{} &       & \multicolumn{1}{r}{} &  \\
		covtype & \multicolumn{1}{r}{0.1521} & \multicolumn{1}{r|}{0.0001} & \multicolumn{1}{r}{0.1520} & \multicolumn{1}{r|}{0.0001} & \multicolumn{1}{r}{0.0506} & \multicolumn{1}{r|}{0.0000} & \multicolumn{1}{r}{0.0507} & \multicolumn{1}{r|}{0.0000} & \multicolumn{1}{r}{0.8583} & \multicolumn{1}{r|}{0.0008} & \multicolumn{1}{r}{0.8867} & \multicolumn{1}{r|}{0.0005} & \multicolumn{1}{r}{0.7052} & \multicolumn{1}{r|}{0.0009} & \multicolumn{1}{r}{0.7820} & \multicolumn{1}{r|}{0.0003} & \multicolumn{1}{r}{} &       & \multicolumn{1}{r}{} &  \bigstrut[b]\\
		\hline
		Wins  & \multicolumn{4}{c|}{3}        & \multicolumn{4}{c|}{12}       & \multicolumn{4}{c|}{9}        & \multicolumn{4}{c|}{14}       & \multicolumn{4}{c|}{5} \bigstrut\\
		\hline
		Mean(MetaIS-IS) & \multicolumn{4}{c|}{-0.0028 (=)} & \multicolumn{4}{c|}{0.0003 (=)} & \multicolumn{4}{c|}{-0.0093 (=)} & \multicolumn{4}{c|}{0.0538 (+)} & \multicolumn{4}{c|}{-0.0144 (-)} \bigstrut\\
		\hline
	\end{tabular}%
	}
	}
	\label{tab:results_AUC1d}%
\end{table}%

An extension of these results is presented in Table \ref{tab:results_AUC} where the entire AUARR is gathered without limitation by the $reduction\_rate$ of the reference instance selection. According to these results, the meta-instance selection outperforms the reference instance selection method in almost all of the cases, only for the Drop3 algorithm the difference is not significantly better. It significantly wins with the reference Drop3 8 out of 19 times.

\begin{table}[htbp]
	\centering
	\caption{Comparison of the reference instance selection and the meta instance selection equivalence. The values represented the area under the accuracy - $reduction\_rate$ curve ($AUARR$). The last two lines summarize the results. The one before the last row indicates the number of times MetaIS was significantly better than the reference model. The last row shows the average difference between the performance of MetaIS and the reference instance selection model over all datasets. The symbols (=), (+), and (-) indicate a statistically significant difference between MetaIS and the reference instance selection. (+) means MetaIS is significantly better than IS, and correspondingly (-) that MetaIS is significantly worse.}
	\Rotatebox{90}{
		\resizebox{
			0.8\textheight
		}{!}{		
	\begin{tabular}{|l|c|c|c|c|c|c|c|c|c|c|c|c|c|c|c|c|c|c|c|c|}
		\hline
		\multicolumn{1}{|c|}{\multirow{3}[2]{*}{Dataset}} & \multicolumn{4}{c|}{HMN-EI}   & \multicolumn{4}{c|}{CCIS}     & \multicolumn{4}{c|}{ENN}      & \multicolumn{4}{c|}{ICF}      & \multicolumn{4}{c|}{Drop3} \bigstrut[t]\\
		& \multicolumn{2}{c|}{IS} & \multicolumn{2}{c|}{MetaIS} & \multicolumn{2}{c|}{IS} & \multicolumn{2}{c|}{MetaIS} & \multicolumn{2}{c|}{IS} & \multicolumn{2}{c|}{MetaIS} & \multicolumn{2}{c|}{IS} & \multicolumn{2}{c|}{MetaIS} & \multicolumn{2}{c|}{IS} & \multicolumn{2}{c|}{MetaIS} \\
		& \multicolumn{1}{l}{mean} & \multicolumn{1}{l|}{std} & \multicolumn{1}{l}{mean} & \multicolumn{1}{l|}{std} & \multicolumn{1}{l}{mean} & \multicolumn{1}{l|}{std} & \multicolumn{1}{l}{mean} & \multicolumn{1}{l|}{std} & \multicolumn{1}{l}{mean} & \multicolumn{1}{l|}{std} & \multicolumn{1}{l}{mean} & \multicolumn{1}{l|}{std} & \multicolumn{1}{l}{mean} & \multicolumn{1}{l|}{std} & \multicolumn{1}{l}{mean} & \multicolumn{1}{l|}{std} & \multicolumn{1}{l}{mean} & \multicolumn{1}{l|}{std} & \multicolumn{1}{l}{mean} & \multicolumn{1}{l|}{std} \bigstrut[b]\\
		\hline
		banana & \multicolumn{1}{r}{0.1023} & \multicolumn{1}{r|}{0.0007} & \multicolumn{1}{r}{0.6350} & \multicolumn{1}{r|}{0.0134} & \multicolumn{1}{r}{0.8249} & \multicolumn{1}{r|}{0.0058} & \multicolumn{1}{r}{0.8263} & \multicolumn{1}{r|}{0.0091} & \multicolumn{1}{r}{0.0989} & \multicolumn{1}{r|}{0.0009} & \multicolumn{1}{r}{0.2839} & \multicolumn{1}{r|}{0.0055} & \multicolumn{1}{r}{0.7447} & \multicolumn{1}{r|}{0.0066} & \multicolumn{1}{r}{0.8610} & \multicolumn{1}{r|}{0.0076} & \multicolumn{1}{r}{0.8096} & \multicolumn{1}{r|}{0.0084} & \multicolumn{1}{r}{0.8385} & \multicolumn{1}{r|}{0.0107} \bigstrut[t]\\
		electricity-normalized & \multicolumn{1}{r}{0.1877} & \multicolumn{1}{r|}{0.0004} & \multicolumn{1}{r}{0.7002} & \multicolumn{1}{r|}{0.0032} & \multicolumn{1}{r}{0.7316} & \multicolumn{1}{r|}{0.0020} & \multicolumn{1}{r}{0.6861} & \multicolumn{1}{r|}{0.0024} & \multicolumn{1}{r}{0.1364} & \multicolumn{1}{r|}{0.0003} & \multicolumn{1}{r}{0.4727} & \multicolumn{1}{r|}{0.0030} & \multicolumn{1}{r}{0.6270} & \multicolumn{1}{r|}{0.0021} & \multicolumn{1}{r}{0.7401} & \multicolumn{1}{r|}{0.0034} & \multicolumn{1}{r}{0.6904} & \multicolumn{1}{r|}{0.0044} & \multicolumn{1}{r}{0.7174} & \multicolumn{1}{r|}{0.0019} \\
		letter & \multicolumn{1}{r}{0.2169} & \multicolumn{1}{r|}{0.0008} & \multicolumn{1}{r}{0.5845} & \multicolumn{1}{r|}{0.0033} & \multicolumn{1}{r}{0.6189} & \multicolumn{1}{r|}{0.0025} & \multicolumn{1}{r}{0.8611} & \multicolumn{1}{r|}{0.0025} & \multicolumn{1}{r}{0.0501} & \multicolumn{1}{r|}{0.0001} & \multicolumn{1}{r}{0.3574} & \multicolumn{1}{r|}{0.0014} & \multicolumn{1}{r}{0.6814} & \multicolumn{1}{r|}{0.0013} & \multicolumn{1}{r}{0.7773} & \multicolumn{1}{r|}{0.0031} & \multicolumn{1}{r}{0.7861} & \multicolumn{1}{r|}{0.0036} & \multicolumn{1}{r}{0.7381} & \multicolumn{1}{r|}{0.0062} \\
		magic & \multicolumn{1}{r}{0.1809} & \multicolumn{1}{r|}{0.0011} & \multicolumn{1}{r}{0.5963} & \multicolumn{1}{r|}{0.0051} & \multicolumn{1}{r}{0.7460} & \multicolumn{1}{r|}{0.0034} & \multicolumn{1}{r}{0.7553} & \multicolumn{1}{r|}{0.0062} & \multicolumn{1}{r}{0.1350} & \multicolumn{1}{r|}{0.0007} & \multicolumn{1}{r}{0.3526} & \multicolumn{1}{r|}{0.0034} & \multicolumn{1}{r}{0.6821} & \multicolumn{1}{r|}{0.0045} & \multicolumn{1}{r}{0.7009} & \multicolumn{1}{r|}{0.0036} & \multicolumn{1}{r}{0.7079} & \multicolumn{1}{r|}{0.0082} & \multicolumn{1}{r}{0.6818} & \multicolumn{1}{r|}{0.0006} \\
		marketing & \multicolumn{1}{r}{0.1438} & \multicolumn{1}{r|}{0.0037} & \multicolumn{1}{r}{0.2914} & \multicolumn{1}{r|}{0.0047} & \multicolumn{1}{r}{0.0574} & \multicolumn{1}{r|}{0.0015} & \multicolumn{1}{r}{0.1400} & \multicolumn{1}{r|}{0.0092} & \multicolumn{1}{r}{0.2068} & \multicolumn{1}{r|}{0.0035} & \multicolumn{1}{r}{0.2771} & \multicolumn{1}{r|}{0.0026} & \multicolumn{1}{r}{0.2295} & \multicolumn{1}{r|}{0.0023} & \multicolumn{1}{r}{0.2897} & \multicolumn{1}{r|}{0.0067} & \multicolumn{1}{r}{0.2365} & \multicolumn{1}{r|}{0.0029} & \multicolumn{1}{r}{0.2849} & \multicolumn{1}{r|}{0.0055} \\
		nursery & \multicolumn{1}{r}{0.2514} & \multicolumn{1}{r|}{0.0041} & \multicolumn{1}{r}{0.7858} & \multicolumn{1}{r|}{0.0212} & \multicolumn{1}{r}{0.7332} & \multicolumn{1}{r|}{0.0110} & \multicolumn{1}{r}{0.8707} & \multicolumn{1}{r|}{0.0121} & \multicolumn{1}{r}{0.0626} & \multicolumn{1}{r|}{0.0014} & \multicolumn{1}{r}{0.4021} & \multicolumn{1}{r|}{0.0305} & \multicolumn{1}{r}{0.7206} & \multicolumn{1}{r|}{0.0103} & \multicolumn{1}{r}{0.8817} & \multicolumn{1}{r|}{0.0125} & \multicolumn{1}{r}{0.7708} & \multicolumn{1}{r|}{0.0234} & \multicolumn{1}{r}{0.8130} & \multicolumn{1}{r|}{0.0103} \\
		optdigits & \multicolumn{1}{r}{0.2029} & \multicolumn{1}{r|}{0.0008} & \multicolumn{1}{r}{0.4241} & \multicolumn{1}{r|}{0.0053} & \multicolumn{1}{r}{0.9078} & \multicolumn{1}{r|}{0.0051} & \multicolumn{1}{r}{0.9427} & \multicolumn{1}{r|}{0.0057} & \multicolumn{1}{r}{0.0222} & \multicolumn{1}{r|}{0.0001} & \multicolumn{1}{r}{0.1528} & \multicolumn{1}{r|}{0.0038} & \multicolumn{1}{r}{0.7681} & \multicolumn{1}{r|}{0.0087} & \multicolumn{1}{r}{0.9113} & \multicolumn{1}{r|}{0.0036} & \multicolumn{1}{r}{0.8650} & \multicolumn{1}{r|}{0.0166} & \multicolumn{1}{r}{0.8549} & \multicolumn{1}{r|}{0.0049} \\
		page-blocks & \multicolumn{1}{r}{0.0317} & \multicolumn{1}{r|}{0.0002} & \multicolumn{1}{r}{0.3769} & \multicolumn{1}{r|}{0.0047} & \multicolumn{1}{r}{0.8691} & \multicolumn{1}{r|}{0.0086} & \multicolumn{1}{r}{0.9369} & \multicolumn{1}{r|}{0.0093} & \multicolumn{1}{r}{0.0321} & \multicolumn{1}{r|}{0.0002} & \multicolumn{1}{r}{0.1135} & \multicolumn{1}{r|}{0.0025} & \multicolumn{1}{r}{0.7205} & \multicolumn{1}{r|}{0.0231} & \multicolumn{1}{r}{0.9063} & \multicolumn{1}{r|}{0.0127} & \multicolumn{1}{r}{0.8849} & \multicolumn{1}{r|}{0.0200} & \multicolumn{1}{r}{0.8563} & \multicolumn{1}{r|}{0.0189} \\
		penbased & \multicolumn{1}{r}{0.1142} & \multicolumn{1}{r|}{0.0002} & \multicolumn{1}{r}{0.3319} & \multicolumn{1}{r|}{0.0070} & \multicolumn{1}{r}{0.9497} & \multicolumn{1}{r|}{0.0051} & \multicolumn{1}{r}{0.9687} & \multicolumn{1}{r|}{0.0011} & \multicolumn{1}{r}{0.0070} & \multicolumn{1}{r|}{0.0000} & \multicolumn{1}{r}{0.0505} & \multicolumn{1}{r|}{0.0012} & \multicolumn{1}{r}{0.8761} & \multicolumn{1}{r|}{0.0015} & \multicolumn{1}{r}{0.9646} & \multicolumn{1}{r|}{0.0025} & \multicolumn{1}{r}{0.9212} & \multicolumn{1}{r|}{0.0075} & \multicolumn{1}{r}{0.9591} & \multicolumn{1}{r|}{0.0038} \\
		phoneme & \multicolumn{1}{r}{0.1521} & \multicolumn{1}{r|}{0.0013} & \multicolumn{1}{r}{0.6084} & \multicolumn{1}{r|}{0.0114} & \multicolumn{1}{r}{0.8142} & \multicolumn{1}{r|}{0.0048} & \multicolumn{1}{r}{0.8242} & \multicolumn{1}{r|}{0.0071} & \multicolumn{1}{r}{0.1007} & \multicolumn{1}{r|}{0.0010} & \multicolumn{1}{r}{0.3822} & \multicolumn{1}{r|}{0.0014} & \multicolumn{1}{r}{0.6592} & \multicolumn{1}{r|}{0.0071} & \multicolumn{1}{r}{0.8033} & \multicolumn{1}{r|}{0.0084} & \multicolumn{1}{r}{0.7458} & \multicolumn{1}{r|}{0.0114} & \multicolumn{1}{r}{0.7651} & \multicolumn{1}{r|}{0.0036} \\
		ring  & \multicolumn{1}{r}{0.5909} & \multicolumn{1}{r|}{0.0018} & \multicolumn{1}{r}{0.7857} & \multicolumn{1}{r|}{0.0095} & \multicolumn{1}{r}{0.6594} & \multicolumn{1}{r|}{0.0024} & \multicolumn{1}{r}{0.7492} & \multicolumn{1}{r|}{0.0084} & \multicolumn{1}{r}{0.2007} & \multicolumn{1}{r|}{0.0010} & \multicolumn{1}{r}{0.3258} & \multicolumn{1}{r|}{0.0018} & \multicolumn{1}{r}{0.6426} & \multicolumn{1}{r|}{0.0052} & \multicolumn{1}{r}{0.6563} & \multicolumn{1}{r|}{0.0100} & \multicolumn{1}{r}{0.6722} & \multicolumn{1}{r|}{0.0046} & \multicolumn{1}{r}{0.7133} & \multicolumn{1}{r|}{0.0035} \\
		satimage & \multicolumn{1}{r}{0.0831} & \multicolumn{1}{r|}{0.0003} & \multicolumn{1}{r}{0.5032} & \multicolumn{1}{r|}{0.0041} & \multicolumn{1}{r}{0.7559} & \multicolumn{1}{r|}{0.0039} & \multicolumn{1}{r}{0.8491} & \multicolumn{1}{r|}{0.0041} & \multicolumn{1}{r}{0.0859} & \multicolumn{1}{r|}{0.0004} & \multicolumn{1}{r}{0.2694} & \multicolumn{1}{r|}{0.0011} & \multicolumn{1}{r}{0.7047} & \multicolumn{1}{r|}{0.0075} & \multicolumn{1}{r}{0.7716} & \multicolumn{1}{r|}{0.0082} & \multicolumn{1}{r}{0.7812} & \multicolumn{1}{r|}{0.0260} & \multicolumn{1}{r}{0.7386} & \multicolumn{1}{r|}{0.0041} \\
		shuttle & \multicolumn{1}{r}{0.0111} & \multicolumn{1}{r|}{0.0000} & \multicolumn{1}{r}{0.5646} & \multicolumn{1}{r|}{0.0075} & \multicolumn{1}{r}{0.9449} & \multicolumn{1}{r|}{0.0158} & \multicolumn{1}{r}{0.9965} & \multicolumn{1}{r|}{0.0005} & \multicolumn{1}{r}{0.0011} & \multicolumn{1}{r|}{0.0000} & \multicolumn{1}{r}{0.0060} & \multicolumn{1}{r|}{0.0003} & \multicolumn{1}{r}{0.7261} & \multicolumn{1}{r|}{0.0198} & \multicolumn{1}{r}{0.9922} & \multicolumn{1}{r|}{0.0012} & \multicolumn{1}{r}{0.9652} & \multicolumn{1}{r|}{0.0239} & \multicolumn{1}{r}{0.9969} & \multicolumn{1}{r|}{0.0005} \\
		spambase & \multicolumn{1}{r}{0.1261} & \multicolumn{1}{r|}{0.0007} & \multicolumn{1}{r}{0.6508} & \multicolumn{1}{r|}{0.0113} & \multicolumn{1}{r}{0.8445} & \multicolumn{1}{r|}{0.0080} & \multicolumn{1}{r}{0.7509} & \multicolumn{1}{r|}{0.0085} & \multicolumn{1}{r}{0.0862} & \multicolumn{1}{r|}{0.0003} & \multicolumn{1}{r}{0.3416} & \multicolumn{1}{r|}{0.0112} & \multicolumn{1}{r}{0.6507} & \multicolumn{1}{r|}{0.0165} & \multicolumn{1}{r}{0.7479} & \multicolumn{1}{r|}{0.0087} & \multicolumn{1}{r}{0.7917} & \multicolumn{1}{r|}{0.0195} & \multicolumn{1}{r}{0.7781} & \multicolumn{1}{r|}{0.0070} \\
		texture & \multicolumn{1}{r}{0.1137} & \multicolumn{1}{r|}{0.0005} & \multicolumn{1}{r}{0.5360} & \multicolumn{1}{r|}{0.0072} & \multicolumn{1}{r}{0.9176} & \multicolumn{1}{r|}{0.0081} & \multicolumn{1}{r}{0.9539} & \multicolumn{1}{r|}{0.0043} & \multicolumn{1}{r}{0.0139} & \multicolumn{1}{r|}{0.0001} & \multicolumn{1}{r}{0.1571} & \multicolumn{1}{r|}{0.0032} & \multicolumn{1}{r}{0.8123} & \multicolumn{1}{r|}{0.0017} & \multicolumn{1}{r}{0.8959} & \multicolumn{1}{r|}{0.0081} & \multicolumn{1}{r}{0.8945} & \multicolumn{1}{r|}{0.0099} & \multicolumn{1}{r}{0.8529} & \multicolumn{1}{r|}{0.0125} \\
		twonorm & \multicolumn{1}{r}{0.1354} & \multicolumn{1}{r|}{0.0007} & \multicolumn{1}{r}{0.5655} & \multicolumn{1}{r|}{0.0068} & \multicolumn{1}{r}{0.9295} & \multicolumn{1}{r|}{0.0033} & \multicolumn{1}{r}{0.9415} & \multicolumn{1}{r|}{0.0041} & \multicolumn{1}{r}{0.0372} & \multicolumn{1}{r|}{0.0002} & \multicolumn{1}{r}{0.1954} & \multicolumn{1}{r|}{0.0044} & \multicolumn{1}{r}{0.8607} & \multicolumn{1}{r|}{0.0158} & \multicolumn{1}{r}{0.9208} & \multicolumn{1}{r|}{0.0039} & \multicolumn{1}{r}{0.8920} & \multicolumn{1}{r|}{0.0056} & \multicolumn{1}{r}{0.9008} & \multicolumn{1}{r|}{0.0026} \\
		codrnaNorm & \multicolumn{1}{r}{0.1296} & \multicolumn{1}{r|}{0.0001} & \multicolumn{1}{r}{0.4754} & \multicolumn{1}{r|}{0.0032} & \multicolumn{1}{r}{0.9303} & \multicolumn{1}{r|}{0.0010} & \multicolumn{1}{r}{0.9456} & \multicolumn{1}{r|}{0.0005} & \multicolumn{1}{r}{0.0459} & \multicolumn{1}{r|}{0.0000} & \multicolumn{1}{r}{0.1660} & \multicolumn{1}{r|}{0.0009} & \multicolumn{1}{r}{0.6869} & \multicolumn{1}{r|}{0.0041} & \multicolumn{1}{r}{0.9278} & \multicolumn{1}{r|}{0.0012} & \multicolumn{1}{r}{} &       & \multicolumn{1}{r}{} &  \\
		covtype & \multicolumn{1}{r}{0.1521} & \multicolumn{1}{r|}{0.0001} & \multicolumn{1}{r}{0.5230} & \multicolumn{1}{r|}{0.0010} & \multicolumn{1}{r}{0.8583} & \multicolumn{1}{r|}{0.0008} & \multicolumn{1}{r}{0.9047} & \multicolumn{1}{r|}{0.0005} & \multicolumn{1}{r}{0.0506} & \multicolumn{1}{r|}{0.0000} & \multicolumn{1}{r}{0.2566} & \multicolumn{1}{r|}{0.0007} & \multicolumn{1}{r}{0.7052} & \multicolumn{1}{r|}{0.0009} & \multicolumn{1}{r}{0.8626} & \multicolumn{1}{r|}{0.0006} & \multicolumn{1}{r}{} &       & \multicolumn{1}{r}{} &  \bigstrut[b]\\
		\hline
		Wins  & \multicolumn{4}{c|}{17}       & \multicolumn{4}{c|}{15}       & \multicolumn{4}{c|}{17}       & \multicolumn{4}{c|}{17}       & \multicolumn{4}{c|}{8} \bigstrut\\
		\hline
		Mean(MetaIS-IS) & \multicolumn{4}{c|}{0.3896 (+)} & \multicolumn{4}{c|}{0.04499  (+)} & \multicolumn{4}{c|}{0.1771 (+)} & \multicolumn{4}{c|}{0.1173 (+)} & \multicolumn{4}{c|}{0.0042 (=)} \bigstrut\\
		\hline
	\end{tabular}%
	}
	}
	\label{tab:results_AUC}%
\end{table}%

\subsection{Execution Time}
\begin{figure}
	\centering
	\begin{subfigure}[b]{0.49\textwidth}
		\centering
		\includegraphics[width=\textwidth]{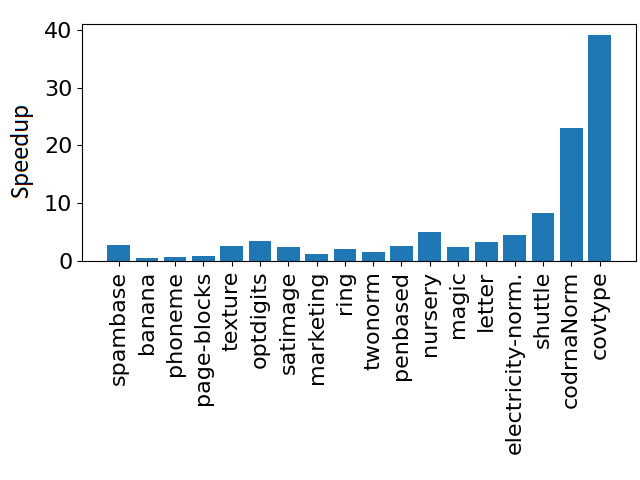}
		\caption{ENN}
		\label{fig:speed_up:enn}
	\end{subfigure}
	\hfill
	\begin{subfigure}[b]{0.49\textwidth}
		\centering
		\includegraphics[width=\textwidth]{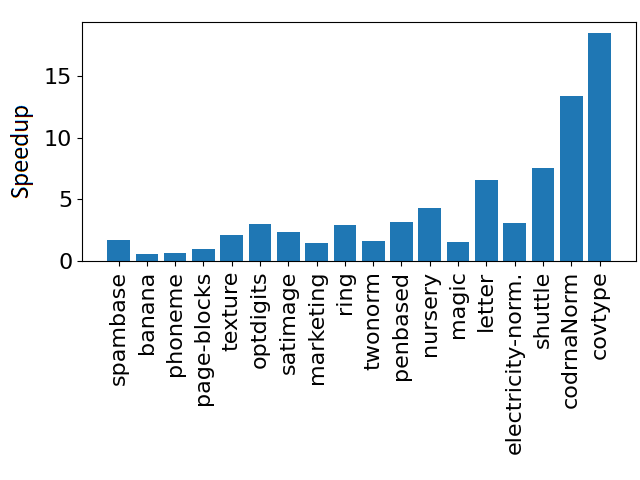}
		\caption{HMN-EI}
		\label{fig:speed_up:hmnei}
	\end{subfigure}
	\begin{subfigure}[b]{0.49\textwidth}
		\centering
		\includegraphics[width=\textwidth]{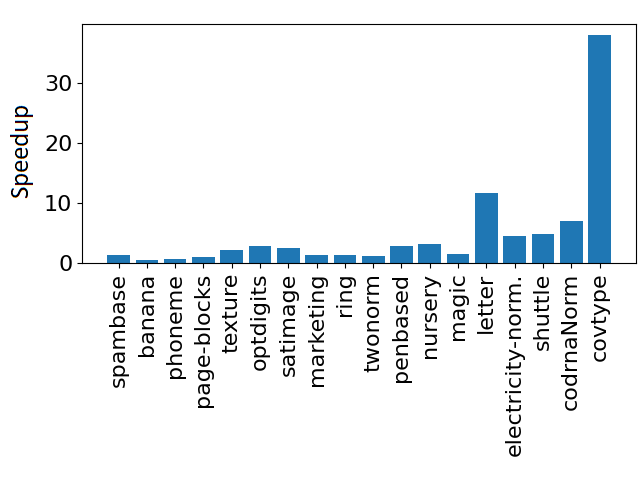}
		\caption{CCIS}
		\label{fig:speed_up:ccis}
	\end{subfigure}
	\hfill
	\begin{subfigure}[b]{0.49\textwidth}
		\centering
		\includegraphics[width=\textwidth]{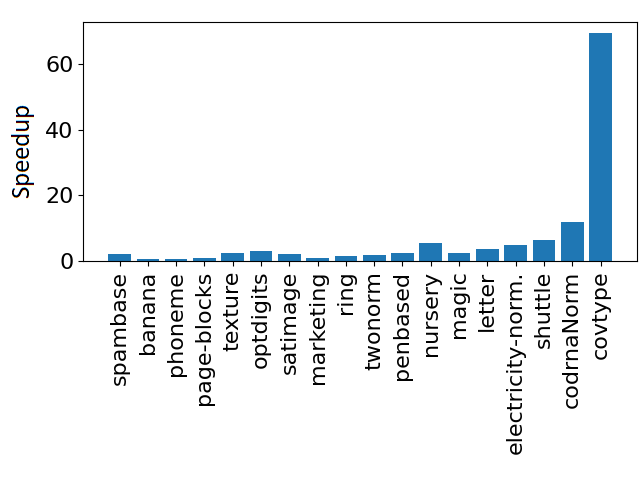}
		\caption{ICF}
		\label{fig:speed_up:icf}
	\end{subfigure}
	\begin{subfigure}[b]{0.49\textwidth}
		\centering
		\includegraphics[width=\textwidth]{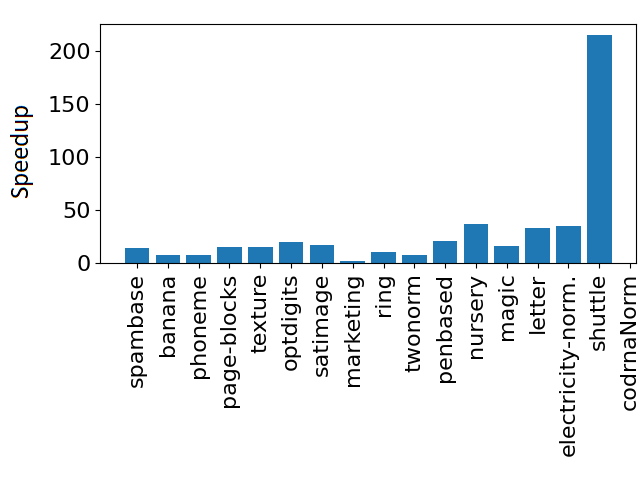}
		\caption{Drop3}
		\label{fig:speed_up:drop3}
	\end{subfigure}	
	\caption{Speedup of the meta instance selection over reference instance selection methods.}
	\label{fig:speed_up}
\end{figure}

The execution time is the utmost benefit of the proposed method. To indicate the benefits and speed up of the meta instance selection over the reference methods, the execution time of the experiments discussed in the previous section were recorded. The execution time was measured only on the instance selection process. For the reference instance selection methods, the execution time was recorded using the build-in RapidMiner execution time property of an operator. For the meta instance selection, two different times can be identified that is the training time and selection time. First, the selection time would be analyzed. The obtained results are presented in Table \ref{tab:execution_time}, and the speed up is shown in Table \ref{tab:speed_up} and visualized in Figure \ref{fig:speed_up}, although it is important to note that the execution time was measured in two different environments - in the case of reference instance selection the algorithms were implemented in Java using the standard java structures and in terms of meta instance selection the algorithms were implemented in Python using numpy and scipy libraries.

\begin{table}[htbp]
	\centering
	\caption{Speedup of the meta instance selection algorithm over the reference instance selection method. The values were obtained as a fraction of the reference instance selection time divided by the time needed by its meta-equivalence.}
	\resizebox{0.85\columnwidth}{!}{
	\begin{tabular}{|l|r|r|r|r|r|r|}
		\hline
		dataset & \multicolumn{1}{l|}{\# samples} & \multicolumn{1}{l|}{HMN-EI [\%]} & \multicolumn{1}{l|}{CCIS [\%]} & \multicolumn{1}{l|}{ENN [\%]} & \multicolumn{1}{l|}{ICF [\%]} & \multicolumn{1}{l|}{Drop3 [\%]} \bigstrut\\
		\hline
		spambase & 4597  & 1.67  & 1.35  & 2.75  & 2.24  & 14.09 \bigstrut[t]\\
		banana & 5300  & 0.56  & 0.42  & 0.50  & 0.51  & 7.63 \\
		phoneme & 5404  & 0.62  & 0.59  & 0.61  & 0.58  & 7.08 \\
		page-blocks & 5472  & 0.95  & 0.97  & 0.91  & 0.90  & 14.65 \\
		texture & 5500  & 2.11  & 2.15  & 2.50  & 2.30  & 15.40 \\
		optdigits & 5620  & 2.98  & 2.74  & 3.53  & 3.10  & 20.17 \\
		satimage & 6435  & 2.31  & 2.44  & 2.48  & 2.10  & 16.68 \\
		marketing & 6876  & 1.47  & 1.28  & 1.25  & 1.01  & 1.55 \\
		ring  & 7400  & 2.90  & 1.29  & 2.02  & 1.42  & 10.46 \\
		twonorm & 7400  & 1.61  & 1.09  & 1.52  & 1.74  & 7.20 \\
		penbased & 10992 & 3.12  & 2.80  & 2.66  & 2.27  & 21.05 \\
		nursery & 12960 & 4.28  & 3.23  & 5.00  & 5.36  & 36.94 \\
		magic & 19020 & 1.57  & 1.51  & 2.49  & 2.37  & 15.64 \\
		letter & 20000 & 6.57  & 11.69 & 3.26  & 3.71  & 33.37 \\
		electricity-norm. & 45312 & 3.08  & 4.46  & 4.49  & 4.88  & 34.90 \\
		shuttle & 57999 & 7.57  & 4.85  & 8.29  & 6.46  & 214.94 \\
		codrnaNorm & 488565 & 13.37 & 7.06  & 22.98 & 11.92 &  \\
		covtype & 581012 & 18.47 & 37.97 & 39.11 & 69.26 &  \bigstrut[b]\\
		\hline
	\end{tabular}%
	}
	\label{tab:speed_up}%
\end{table}%

The analysis of the results presented in Table \ref{tab:speed_up} and visualized in Figure \ref{fig:speed_up} indicates a significant speedup of the meta instance selection algorithm that reaches 219 for the Drop3 algorithm. 
Moreover, when analyzing Table \ref{tab:execution_time} it can observed that the execution time of the meta instance selection is independent of the reference algorithm. It is fixed and depends on the meta-feature extraction process and execution of the meta-classifier. This property is essential for the use of the proposed solution together with the very complex and slow instance selection algorithms with high computational complexity.

\begin{table}[htbp]
	\centering
	\caption{Comparison of the execution time of the reference instance selection method and the meta instance selection equivalence. The values are represented in seconds. For the codrnaNorm and covtype the values for the Drop3 are missing since after 6 days the reference algorithm didn't converge.}
	\resizebox{\columnwidth}{!}{
	\begin{tabular}{|l|r|rr|rr|rr|rr|rr|}
		\hline
		\multicolumn{1}{|c|}{\multirow{2}[2]{*}{dataset}} & \multicolumn{1}{c|}{
			\multirow{2}[2]{*}{\# \newline{}samples}
			} & \multicolumn{2}{c|}{HMEI [s]} & \multicolumn{2}{c|}{CCIS [s]} & \multicolumn{2}{c|}{ENN [s]} & \multicolumn{2}{c|}{ICF [s]} & \multicolumn{2}{c|}{Drop3 [s]} \bigstrut[t]\\
		&       & \multicolumn{1}{l}{Ref} & \multicolumn{1}{l|}{Meta} & \multicolumn{1}{l}{Ref} & \multicolumn{1}{l|}{Meta} & \multicolumn{1}{l}{Ref} & \multicolumn{1}{l|}{Meta} & \multicolumn{1}{l}{Ref} & \multicolumn{1}{l|}{Meta} & \multicolumn{1}{l}{Ref} & \multicolumn{1}{l|}{Meta} \bigstrut[b]\\
		\hline
		spambase & 4597  & 4     & 3     & 3     & 3     & 7     & 2     & 6     & 3     & 35    & 2 \bigstrut[t]\\
		banana & 5300  & 1     & 3     & 1     & 3     & 1     & 3     & 1     & 3     & 19    & 2 \\
		phoneme & 5404  & 2     & 3     & 2     & 3     & 2     & 3     & 2     & 3     & 20    & 3 \\
		page-blocks & 5472  & 3     & 3     & 3     & 3     & 2     & 3     & 3     & 3     & 40    & 3 \\
		texture & 5500  & 6     & 3     & 6     & 3     & 7     & 3     & 7     & 3     & 42    & 3 \\
		optdigits & 5620  & 8     & 3     & 8     & 3     & 9     & 3     & 9     & 3     & 52    & 3 \\
		satimage & 6435  & 8     & 3     & 8     & 3     & 8     & 3     & 7     & 3     & 52    & 3 \\
		marketing & 6876  & 5     & 3     & 4     & 3     & 4     & 3     & 4     & 3     & 5     & 3 \\
		ring  & 7400  & 11    & 4     & 5     & 3     & 7     & 3     & 5     & 4     & 36    & 3 \\
		twonorm & 7400  & 6     & 4     & 4     & 3     & 5     & 3     & 6     & 4     & 24    & 3 \\
		penbased & 10992 & 15    & 5     & 14    & 5     & 13    & 5     & 11    & 5     & 99    & 5 \\
		nursery & 12960 & 29    & 7     & 21    & 7     & 33    & 7     & 36    & 7     & 241   & 7 \\
		magic & 19020 & 15    & 10    & 14    & 9     & 23    & 9     & 22    & 9     & 147   & 9 \\
		letter & 20000 & 66    & 10    & 114   & 10    & 30    & 9     & 37    & 10    & 320   & 10 \\
		electricity-norm. & 45312 & 86    & 28    & 120   & 27    & 127   & 28    & 132   & 27    & 950   & 27 \\
		shuttle & 57999 & 268   & 35    & 171   & 35    & 288   & 35    & 228   & 35    & 7632  & 36 \\
		codrnaNorm & 488565 & 18393 & 1375  & 9396  & 1330  & 29637 & 1290  & 15469 & 1298  &       &  \\
		covtype & 581012 & 55658 & 3013  & 109719 & 2890  & 113242 & 2895  & 194440 & 2807  &       &  \bigstrut[b]\\
		\hline
	\end{tabular}%
	}
	\label{tab:execution_time}%
\end{table}%

The meta instance selection training time should also be considered since the meta classifier needs to be trained. This process is executed once ergo it does not influence the execution time of the selection phase. The training time may be estimated by the addition of the execution time of the selection phase of all datasets that were presented in Table \ref{tab:execution_time}. It may be considered as the reasonable estimation since considering the computational complexity of the random forest classifier and the meta-feature extraction time, this second phase determines the execution time.

\subsection{Assessment of the Meta-Classifier Performance}
A substantial factor influencing the quality of the meta instance selection is the evaluation and selection of an appropriate meta-classifier. In the conducted studies, models based on the random forest algorithm were adopted as the meta-classifier, mainly due to its scalability as a function of the number of training vectors. However, the problem is the imbalanced distribution of class labels in the meta training set. It appears when the meta training set is labeled by the results of the reference instance selection method with a high value of $reduction\_rate$. For example, if the reference instance selection method is characterized by $reduction\_rate=90\%$, the label's distribution will also share identical properties. This introduces certain limitations for the constructed classifier. Therefore, a comparison of the standard random forest with the Balanced Random Forest algorithm \cite{JMLR:v18:16-365} is made below. A Balanced Random Forest differs from a classical Random Forest in that it creates a bootstrap sample from the minority class and then samples with replacement an equal number of instances from the majority class, therefore it is also faster during training.

The obtained results for each of the reference instance selection models are presented in Table \ref{tab:meta_classifier_res} along with the unbalanced rate calculated as the ratio of the number of samples of the positive class to the total number of samples in the meta training set. For the comparison, three different measures of prediction accuracy were used: average accuracy, average balanced accuracy, and the area under the ROC curve - the so-called AUC coefficient. Since, as mentioned above, the analyzed data set is imbalanced, the meta classifier does not return the binary labels, rather the values of the decision function or class probabilities are returned.

The imbalanced classification problem influences the choice of the model evaluation measure, which has a significant impact on the comparison of results. Therefore, the AUC was used in the study as a measure independent of the acceptance threshold. It allows for a fair comparison of the quality of the prediction performance. The balanced accuracy measure is particularly important since it naturally returns the prediction results taking into account the class label distribution and assures that the model with a higher balanced accuracy rate treats both classes equally. Thus, it naturally offers a proper assessment of the model's accuracy. The ordinary assessment of the prediction accuracy was included to indicate a reference value.

\begin{table}[htbp]
	\centering
	\caption{Comparison of the performance of Balanced Random Forest and classical Random Forest in application to the design of a meta-classifier.}
	\resizebox{\columnwidth}{!}{	
	\begin{tabular}{|c|l|c|rr|rr|rr|}
		\hline
		\multirow{2}[2]{*}{Model} & \multicolumn{1}{c|}{\multirow{2}[2]{*}{Classifier}} & \multicolumn{1}{p{4.855em}|}{Imbalnced} & \multicolumn{2}{c|}{AUC [\%]} & \multicolumn{2}{c|}{Balanced Acc [\%]} & \multicolumn{2}{c|}{Accuracy [\%]} \bigstrut[t]\\
		&       &  rate [\%] & \multicolumn{1}{c}{Mean} & \multicolumn{1}{c|}{Std} & \multicolumn{1}{c}{Mean} & \multicolumn{1}{c|}{Std} & \multicolumn{1}{c}{Mean} & \multicolumn{1}{c|}{Std} \bigstrut[b]\\
		\hline
		\multirow{2}[2]{*}{HMN-EI} & RF    & \multirow{2}[2]{*}{0.8059} & 0.9872 & 0.0010 & 0.9415 & 0.0020 & 0.9678 & 0.0010 \bigstrut[t]\\
		& Bal. RF &       & 0.9834 & 0.0009 & 0.9351 & 0.0018 & 0.9227 & 0.0013 \bigstrut[b]\\
		\hline
		\multirow{2}[2]{*}{ENN} & RF    & \multirow{2}[2]{*}{0.8919} & 0.9929 & 0.0003 & 0.9288 & 0.0029 & 0.9748 & 0.0009 \bigstrut[t]\\
		& Bal. RF &       & 0.9925 & 0.0003 & 0.9598 & 0.0014 & 0.9490 & 0.0010 \bigstrut[b]\\
		\hline
		\multirow{2}[2]{*}{ICF} & RF    & \multirow{2}[2]{*}{0.1385} & 0.9146 & 0.0024 & 0.7503 & 0.0017 & 0.9155 & 0.0009 \bigstrut[t]\\
		& Bal. RF &       & 0.9119 & 0.0020 & 0.8349 & 0.0031 & 0.8386 & 0.0013 \bigstrut[b]\\
		\hline
		\multirow{2}[2]{*}{CCIS} & RF    & \multirow{2}[2]{*}{0.1360} & 0.9690 & 0.0011 & 0.8909 & 0.0027 & 0.9659 & 0.0007 \bigstrut[t]\\
		& Bal. RF &       & 0.9581 & 0.0012 & 0.8862 & 0.0025 & 0.8724 & 0.0020 \bigstrut[b]\\
		\hline
		\multirow{2}[2]{*}{Drop3} & RF    & \multirow{2}[2]{*}{0.0686} & 0.9711 & 0.0018 & 0.7888 & 0.0038 & 0.9626 & 0.0009 \bigstrut[t]\\
		& Bal. RF &       & 0.9695 & 0.0015 & 0.9121 & 0.0030 & 0.8857 & 0.0020 \bigstrut[b]\\
		\hline
	\end{tabular}%
	}
	\label{tab:meta_classifier_res}%
\end{table}%

The obtained results indicate that in terms of AUC, models based on Balanced Random Forest and classical Random Forests are similar, however, statistical tests always indicated a significant statistical difference (Welch's T-test for $\alpha=0.05$ was used), indicating the dominance of the Random Forest algorithm. However, the results obtained for balanced accuracy show the opposite results, where, except for the HMN-EI and CCIS algorithms, the Balanced Random Forest algorithm always obtained better results, and the differences reached over 12\% in favor of the Balanced Random Forest algorithm (for HMN-EI and CCIS the differences were small at the level of tenths of a percent). The opposite situation occurred in the case of ordinary prediction accuracy, where the differences reached 9\% in favor of the Random Forest algorithm. In the conducted research, however, it was decided to use the Balanced Random Forest algorithm, because it provides more balanced results for the minority class, which makes the selection of the acceptance threshold $\Theta$ more intuitive, and values at the level of $\Theta=0.5$ are similar to the results obtained for the reference algorithms. On the other hand, for the classical Random Forest algorithm, the selection of the parameter $\Theta$ is a separate issue, which is specific for each basic instance selection algorithm and could be a subject of extended research.

\subsection{Meta-Features Importance Analysis}
The another important element influencing the quality of the obtained results is the selection of the feature space, albeit there is the associated issue as individual meta-features are strongly correlated with each other. This correlation can be visible preeminently  for the same function extracting a meta-feature for different values of the k parameter. Hence in the process of analyzing the significance of features, grouping features with similar properties and presenting their sum was performed.

\begin{figure}
	\centering
	\begin{subfigure}[b]{0.49\textwidth}
		\centering
		\includegraphics[width=\textwidth]{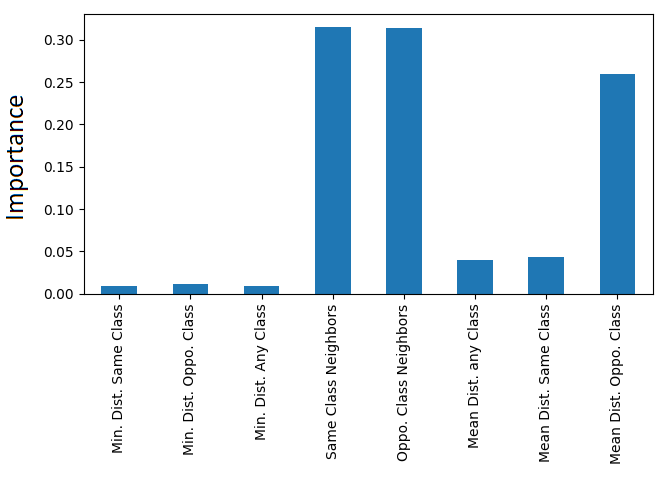}
		\caption{ENN}
		\label{fig:feat_imp:enn}
	\end{subfigure}
	\hfill
	\begin{subfigure}[b]{0.49\textwidth}
		\centering
		\includegraphics[width=\textwidth]{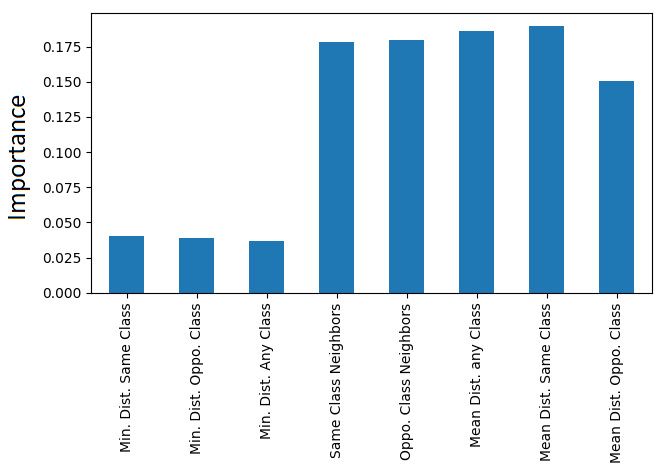}
		\caption{HMN-EI}
		\label{fig:feat_imp:hmnei}
	\end{subfigure}
	\begin{subfigure}[b]{0.49\textwidth}
		\centering
		\includegraphics[width=\textwidth]{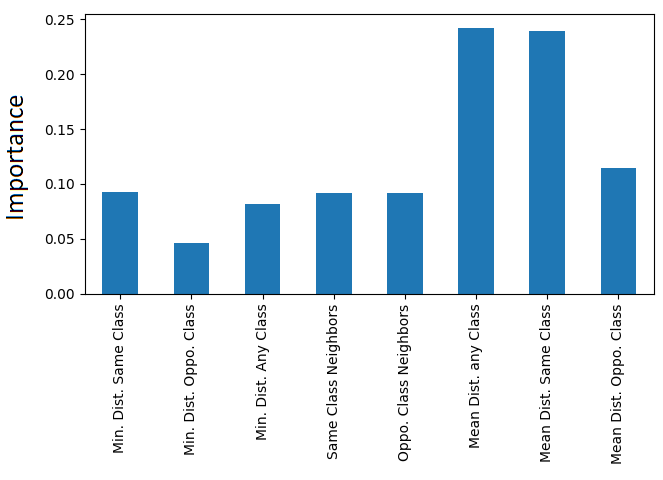}
		\caption{CCIS}
		\label{fig:feat_imp:ccis}
	\end{subfigure}
	\hfill
	\begin{subfigure}[b]{0.49\textwidth}
		\centering
		\includegraphics[width=\textwidth]{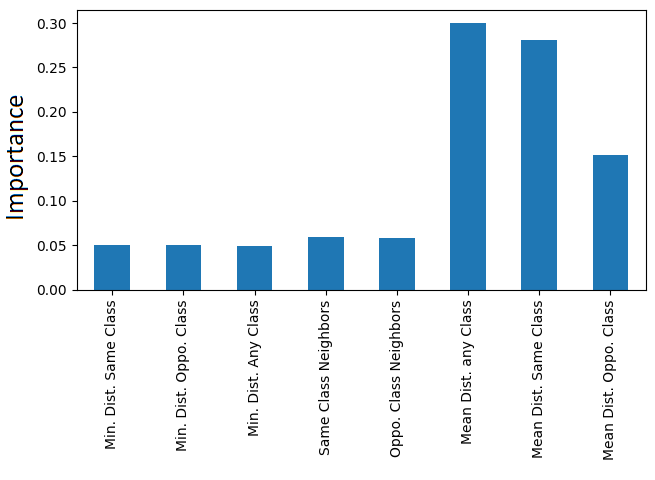}
		\caption{ICF}
		\label{fig:feat_imp:icf}
	\end{subfigure}
	\begin{subfigure}[b]{0.49\textwidth}
		\centering
		\includegraphics[width=\textwidth]{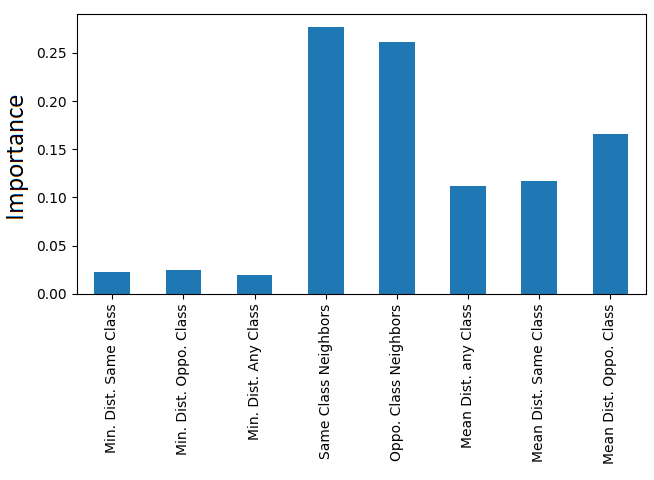}
		\caption{Drop3}
		\label{fig:feat_imp:drop3}
	\end{subfigure}
	\caption{Sum of feature importances obtained after grouping by the feature type.}
	\label{fig:feat_imp}	
\end{figure}

\begin{figure}
	\centering
	\begin{subfigure}[b]{0.49\textwidth}
		\centering
		\includegraphics[width=\textwidth]{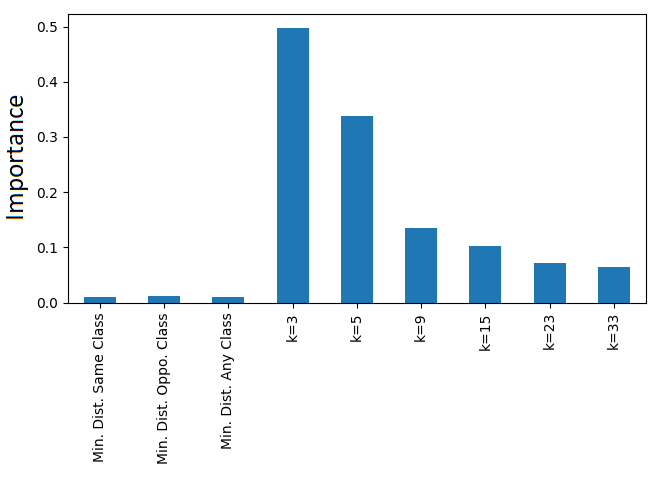}
		\caption{ENN}
		\label{fig:feat_imp_k:enn}
	\end{subfigure}
	\hfill
	\begin{subfigure}[b]{0.49\textwidth}
		\centering
		\includegraphics[width=\textwidth]{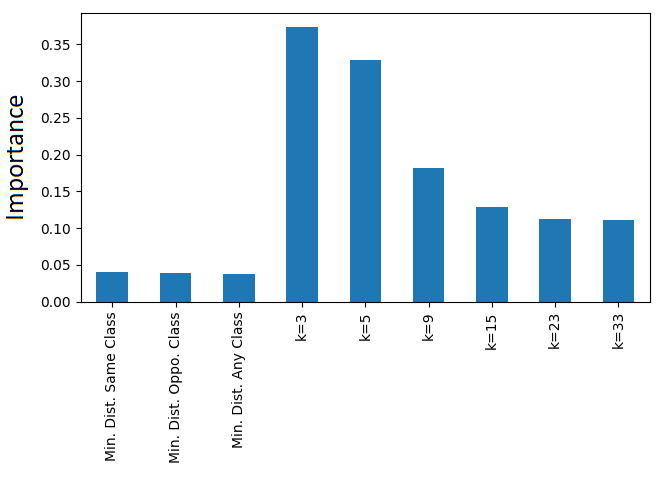}
		\caption{HMN-EI}
		\label{fig:feat_imp_k:hmnei}
	\end{subfigure}
	\begin{subfigure}[b]{0.49\textwidth}
		\centering
		\includegraphics[width=\textwidth]{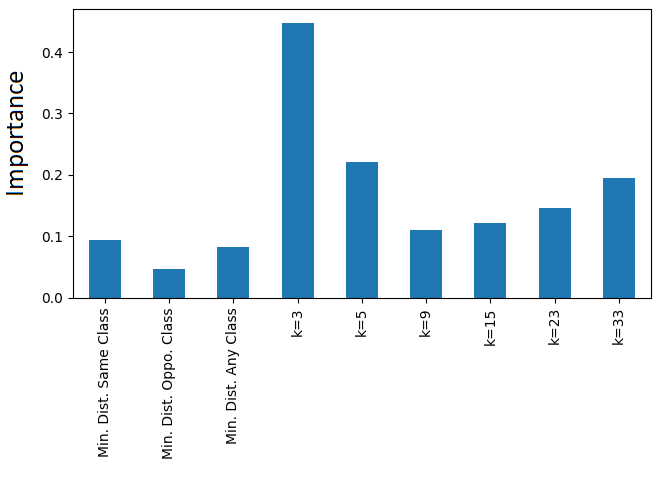}
		\caption{CCIS}
		\label{fig:feat_imp_k:ccis}
	\end{subfigure}
	\hfill
	\begin{subfigure}[b]{0.49\textwidth}
		\centering
		\includegraphics[width=\textwidth]{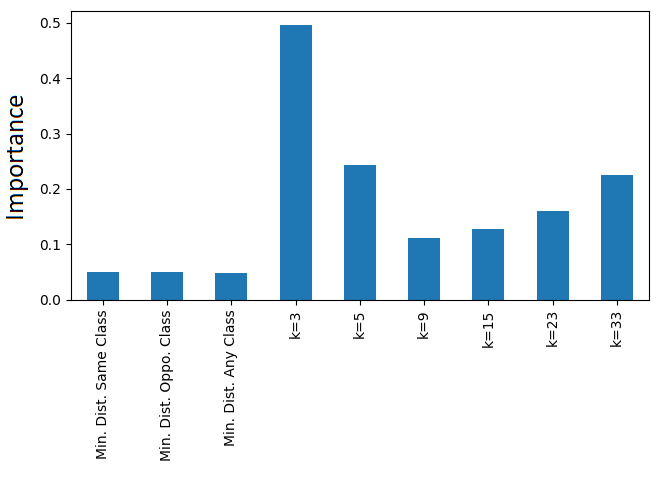}
		\caption{ICF}
		\label{fig:feat_imp_k:icf}
	\end{subfigure}
	\begin{subfigure}[b]{0.49\textwidth}
		\centering
		\includegraphics[width=\textwidth]{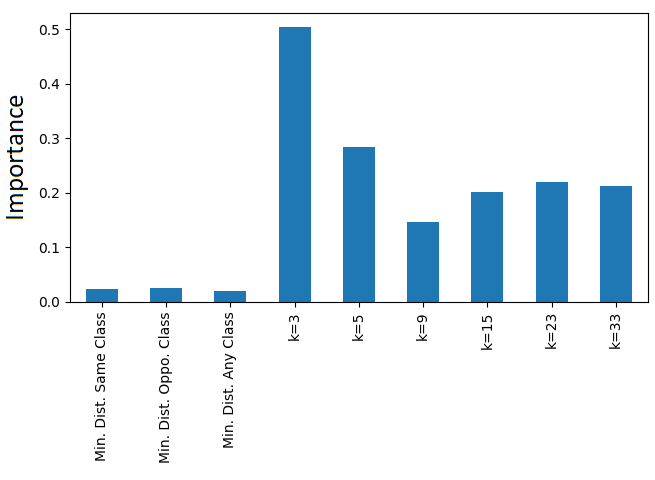}
		\caption{Drop3}
		\label{fig:feat_imp_k:drop3}
	\end{subfigure}	
	\caption{Sum of feature importances obtained after grouping by the feature type.}
	\label{fig:feat_imp_k}
\end{figure}

The mean decrease in impurity (MDI) method \cite{louppe2013understanding} was used to assess the quality of attributes, which is naturally supported by tree-based algorithms such as Random Forest. The grouping of features was performed based on two aggregation methods. First the meta-features were grouped by the feature types and summed the MDI coefficients for different values of $k$ and the second aggregation method was based on the value of $k$, then the coefficients for different feature types and the same value of $k$ were summed. The obtained results are presented graphically in Figure \ref{fig:feat_imp} and Figure \ref{fig:feat_imp_k}, respectively.



The obtained results indicate a lack of uniformity of results between the individual meta-models. In terms of attribute grouping by feature types, it can be observed that for the ENN algorithm, the most important features are the number of instances from the same and opposite class among $k$ nearest neighbors. At the same time, the meta-classifier also uses information about the average distance to the opposite class when making decisions. This result is understandable due to the way the ENN algorithm works. In the case of the HMN-EI algorithm, the most important features are a set of meta-attributes describing the statistics of the occurrence of the nearest neighbors belonging to the same and opposite class as well as average distances. Attributes describing the distances to the nearest samples from the opposite class and the same class are ignored here. The above corresponds to the statistics analyzed by the Hit Miss Network, which is the basis of the HME-EI algorithm. For the CCIS algorithm, the most important attribute is also the average distance to the nearest neighbors, including those from the same class, but the remaining attributes of the dataset are also relatively important especially when compared to the other methods. The attribute importances for the ICF algorithm present a very similar characteristic to CCIS. In this case, a criterion such as Caverage can be relatively easily identified, but the Reachable criterion must be estimated based on the remaining features. In the case of the Drop3 algorithm, the attribute importances are similar to the ENN algorithm, which is justified, because the Drop3 algorithm starts with the ENN algorithm, followed by the proper instance selection procedure. At the same time, due to the need to arrange the removed vectors in order defined by the distance to the nearest enemy, the following most important attribute is the average distance to the nearest instance from the opposite class.

In the case of grouping attribute importance by parameter $k$, it can be observed that usually the most important feature is $k=3$. This is apprehensible since each of these algorithms was initialized specifically for $k=3$. For the ENN and HMN-EI algorithms, the decrease in attribute importance for increasing values of $k$ may be observed. 
On the other hand, for the CCIS, ICF and Drop3 algorithms, attribute importance for subsequent values of $k$ reaches a minimum for $k=9$, after which there is a gradual increase in attribute importance for higher $k$. The above results from the iterative nature of the algorithms. These algorithms complete the selection process in subsequent iterations, in this case increasing the number of $k$ allows for a better estimation of the selection process results.


\section{Conclusions} \label{sec:conclusions}
This paper presents the results of research related to the training set selection system using meta-features and a meta-classifier. The developed method consists of converting classical instance selection methods into the binary classification problem where each training sample is classified as "to be removed"/"to be kept" using any classifier operating in the meta-feature space. These meta-features represent samples of any data set in a common feature space that is extracted from the nearest neighbors graph. As shown in the paper, this methodology allows for significant acceleration of the calculations, due to the fact that results may be obtained with a single-pass analysis of each of the vectors, without the need to implement an iterative process. It was also shown that the proposed method is generic and can be used with any instance selection method utilizing a nearest-neighbor graph for instance assessment. Furthermore, the developed solution allows for dynamic adjustment of the size of the resulting training set, because the value returned from the system represents the classifier's decision function or the value of the probability of belonging to the "to be removed" class. Then, the user can achieve the desired training set compression by setting appropriate acceptance threshold $\Theta$. 

When building a meta-classifier and dealing with imbalanced classification the Balanced Random Forest classifier is a better choice than the classical Random Forest, because it 
facilitates the selection of the appropriate value of the $\Theta$ parameter.

Finally, the results of meta-features importance show that different methods have particular preferences and focus on various characteristics of the NNG. This opens a room for feature research. For example, one of the options is the ensemble model. Since the the meta-classifier for particular instance selection methods focuses on different feature subsets the obtained ensemble should have a guaranteed diversity. Due to the breadth of this problem, the above will go beyond the scope of this study thus it is not presented within the scope of the article.

\section[*]{Acknowledgements}
Publication supported by the Excellence Initiative – Research University program implemented at the Silesian University of Technology, year 2024

\end{document}